\pdfoutput=1

\documentclass[11pt]{article}

\usepackage[]{acl}
\usepackage{times}
\usepackage{latexsym}

\usepackage[T1]{fontenc}

\usepackage[utf8]{inputenc}

\usepackage{microtype}

\usepackage{inconsolata}
\usepackage[caption=false, font=footnotesize]{subfig}
\usepackage{graphicx} 
\usepackage{multirow}

\DeclareRobustCommand{\company}{Megagon Labs}
\newcommand{\stitle}[1]{\vspace{0.2em}\noindent\textbf{#1}}
\newcommand{\hide}[1]{}
\newcommand{\eg}{{\itshape e.g.}, }
\newcommand{\ie}{{\itshape i.e.}, }
\newcommand{\code}[1]{\texttt{\small #1}}

\newcommand{\papertext}[1]{}

\title{Characterizing Large Language Models as Rationalizers of Knowledge-intensive Tasks}

\author{Aditi Mishra\\ Arizona State University \\ \texttt{amishr45@asu.edu} \\\And
  Sajjadur Rahman \\ Megagon Labs, USA \\ \texttt{sajjadur@megagon.ai} \\\And
  Hannah Kim \\ Megagon Labs, USA \\ \texttt{hannah@megagon.ai} \\ 
  \AND 
  Kushan Mitra \\
  Megagon Labs, USA \\
  \texttt{kushan@megagon.ai} \\\And
  Estevam Hruschka \\
  Megagon Labs, USA \\
  \texttt{estevam@megagon.ai} \\}

\begin{document}
\maketitle
\begin{abstract}
Large language models (LLMs) are proficient at generating fluent text with minimal task-specific supervision. However, their ability to generate rationales for knowledge-intensive tasks (KITs) remains under-explored. Generating rationales for KIT solutions, such as commonsense multiple-choice QA, requires external knowledge to support predictions and refute alternate options. In this work, we consider the task of generating knowledge-guided rationalization of KIT model predictions in natural language by using expert-written examples in a few-shot manner. Surprisingly, crowd-workers preferred knowledge-grounded rationales over human-written rationalizations, citing their factuality, sufficiency, and convincingness. Although LLM-generated rationales were preferable, further improvements in conciseness and novelty are required. Moreover, we demonstrate how rationalizing incorrect model predictions erodes humans' trust in LLM-generated rationales. Motivated by these observations, we create a two-stage pipeline to review KIT model predictions before rationalization, enabling trustworthy rationale generation.
\end{abstract}

\section{Introduction}
\label{sec:introduction}
In recent years, generating \emph{rationales} (\ie free-text explanations) of natural language understanding tasks 
has been increasingly explored in the field of explainable NLP.
Such rationales --- while less functionally grounded, \ie they may not entirely reflect the model’s behavior --- provide an effective interface to interpretably communicate model decisions to end-users~\cite{hendricks2016generating, camburu2018snli, madsen2022post, gurrapu2023rationalization}.
Generating these rationales via direct supervision~\cite{ehsan2018rationalization, narang2020wt5} or fine-tuning~\cite{aggarwaletal2021ecqa,rei2022comet} 
requires the collection of high-quality human-authored rationales, which can be expensive, difficult to standardize, and cannot be generalized to all domains~\cite{wiegreffe2021teach, tan2021diversity}.
Recent work~\cite{wiegreffe-etal-2022-reframing} showcases that large language model (LLM) generated rationales, obtained via few-shot in-context learning~\cite{radford2019language, brown2020language,huang2023can}, alleviate these challenges while showcasing surprising effectiveness over crowdsourced rationales on dimensions such as human preference. However, the suitability of LLM-generated rationales for knowledge-intensive tasks (KITs) such as commonsense question answering (CSQA~\cite{csqa}) and open book question answering (OBQA~\cite{obqa}) remains underexplored. 

\begin{figure}
    \centering
    \includegraphics[width=\linewidth]{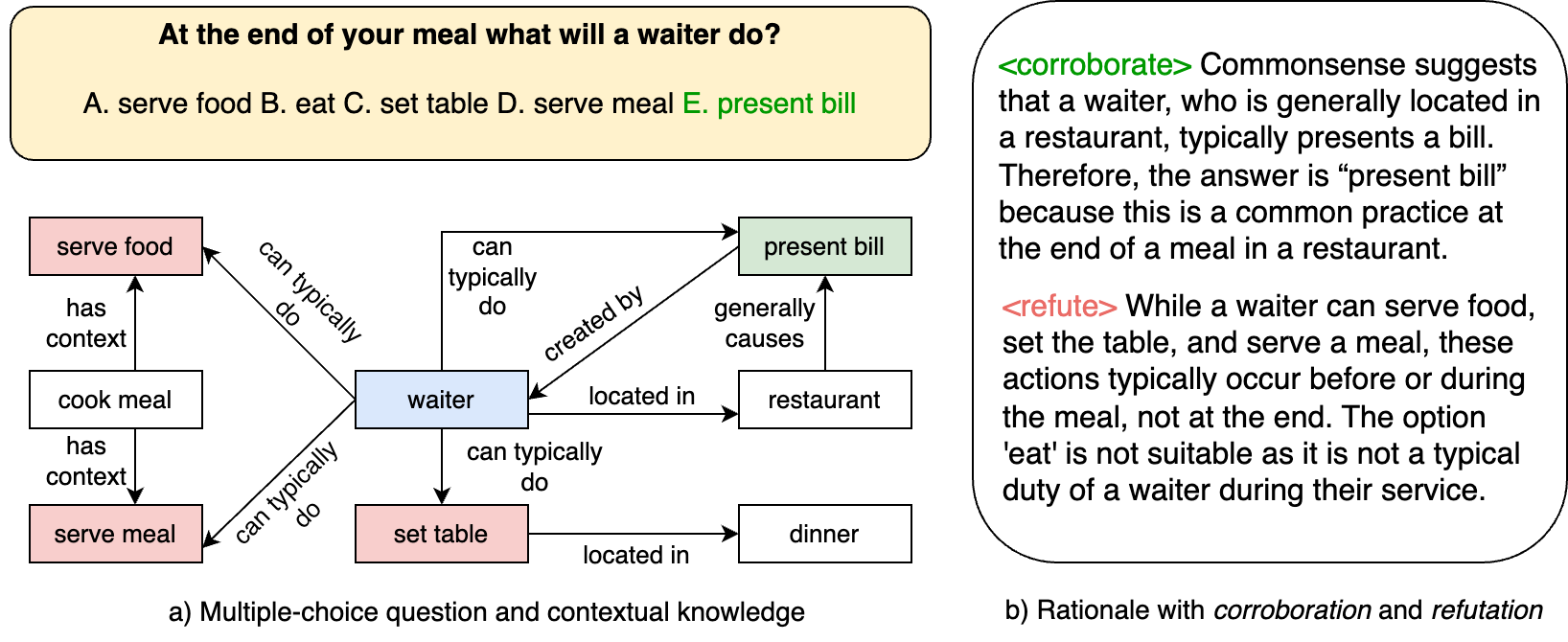}
    \caption{a) A commonsense question with multiple choices and knowledge extracted from ConceptNet and b) proposed LLM-generated rationale corroborating the selected answer and refuting the other choices.}
    \label{fig:kit-example}
\vspace{-10pt}
\end{figure}

Firstly, KITs such as CSQA and OBQA are framed as multiple-choice questions, requiring models to select one answer from several choices (see Figure~\ref{fig:kit-example}a). Therefore, a corresponding well-formed rationale is required to be (a) comprehensive, \ie state facts that are not present in the question but are essential
for rationalization, and (b) refutation complete, \ie rationalize why the rest of the choices are incorrect or not best suited as the answer~\cite{aggarwaletal2021ecqa}. We show an example of such a rationale in Figure~\ref{fig:kit-example}b. However, existing LLM-generated rationales have only been evaluated on their corroboration capabilities~\cite{wiegreffe-etal-2022-reframing}. Secondly, these rationales are abstractive~\cite{gurrapu2023rationalization}, lacking any grounding on external knowledge sources crucial for solving the task --- KIT models designed for CSQA and OBQA~\cite{mhgrn, qagnn, dragon} refer to external sources such as ConceptNet~\cite{speer2017conceptnet} (see Figure~\ref{fig:kit-example}a). 
Finally, KIT models may predict incorrectly --- faithfully rationalizing such mistakes may erode the end-user's trust in the generated rationales. Existing approaches omit the incorrect prediction confounder and evaluate only rationales of correct predictions. 



To address these gaps, we generate knowledge-grounded and refutation complete
rationales --- similar to Figure~\ref{fig:kit-example}b --- via few-shot prompting of LLMs. Examples in the prompt are enriched with relevant knowledge retrieved from external sources to condition the rationale generation on facts.
Two human subjects studies demonstrate that, more often than not, humans prefer LLM-generated
rationales to crowdsourced rationales in existing datasets, citing their factuality, sufficiency, and convincing refutation. 
Follow-up fine-grained analysis 
reveals that LLM-generated rationales
still have significant room for improvement along
dimensions such as \emph{insightfulness} (\ie providing new information) and \emph{redundancy} (\ie avoiding repetitive text).
In another study, motivated by existing literature on trust in explainable AI~\cite{hoffman2018metrics, stites2021sage},
we find that faithful rationalization of incorrect model predictions 
degrades humans' trust in the generated rationales. 
We propose a review-then-rationalize pipeline  
that helps intervene up to $71\%$ of the incorrect predictions. We will publicly release our code and data.


\section{Knowledge-enhanced Rationalization}
\label{sec:pipeline}

\begin{figure}[!htb]
    \centering
    \includegraphics[width=\linewidth]{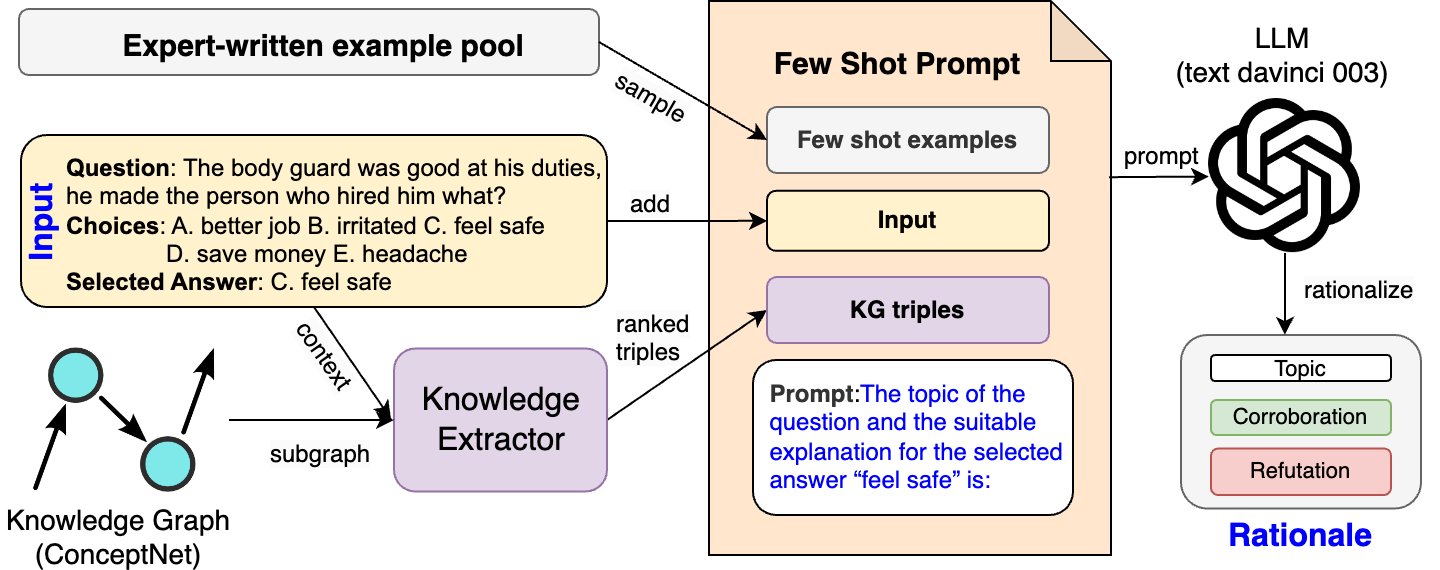}
    \caption{Given an Input (\ie QA and model prediction), an LLM is prompted to generate a rationale with few-shot examples sampled from an expert-written pool.}
    \label{fig:rationalizer}
\vspace{-10pt}
\end{figure}

KIT models such as MHGRN~\cite{mhgrn}, QAGNN~\cite{qagnn}, and Dragon~\cite{dragon} combine language model and knowledge graph representations to solve complex tasks such as commonsense QA~\cite{csqa}. We aim to generate rationales that corroborate the KIT model's prediction with additional relevant facts while refuting the other choices (see Figure~\ref{fig:kit-example}.) Our approach is similar to existing knowledge-guided strategies with LLMs~\cite{peng2023check, lazaridou2022internet, zhao2023verify, mei2023foveate}, which have shown to be effective in text generation.
To guide the generation of these rationale components, \ie corroboration and refutation, we retrieve facts concerning the knowledge-intensive task --- \eg questions and choices in CSQA and OBQA --- from a knowledge graph such as ConceptNet~\cite{speer2017conceptnet}. We then prompt an LLM to rationalize the prediction via conditioning on the provided knowledge. Figure~\ref{fig:rationalizer} outlines the rationalization process given an \emph{input}, \ie question, choices, and model prediction.

\stitle{Knowledge retrieval.} Given the input and an external knowledge-graph such as ConceptNet~\cite{speer2017conceptnet},  
a \emph{Knowledge Extractor} retrieves facts related to the concepts within the question and the choices. We employ a heuristic edge extraction method called QAGNN~\cite{qagnn} to retrieve the relevant facts, \ie triples extracted from the graph. We transform the triples into natural language statements and include those in the prompt used to query the LLM. The retrieved knowledge can be large and violate the maximum allowable tokens in a prompt. For example, the GPT-3.5 \code{text-davinci-003}\footnote{\url{https://platform.openai.com/docs/models/}} supports a maximum of 4,097 tokens per prompt. 
Therefore, within the Knowledge Extractor, we employ RoBERTa~\cite{liu2019roberta} to select top-$k$ ($k=5$) facts given the question and a choice. Similar strategies were adopted in existing work on hybrid sub-graph extraction~\cite{wang-etal-2020-connecting, yan-etal-2021-learning}. 

\begin{figure}
    \centering
    \includegraphics[width=0.7\linewidth]{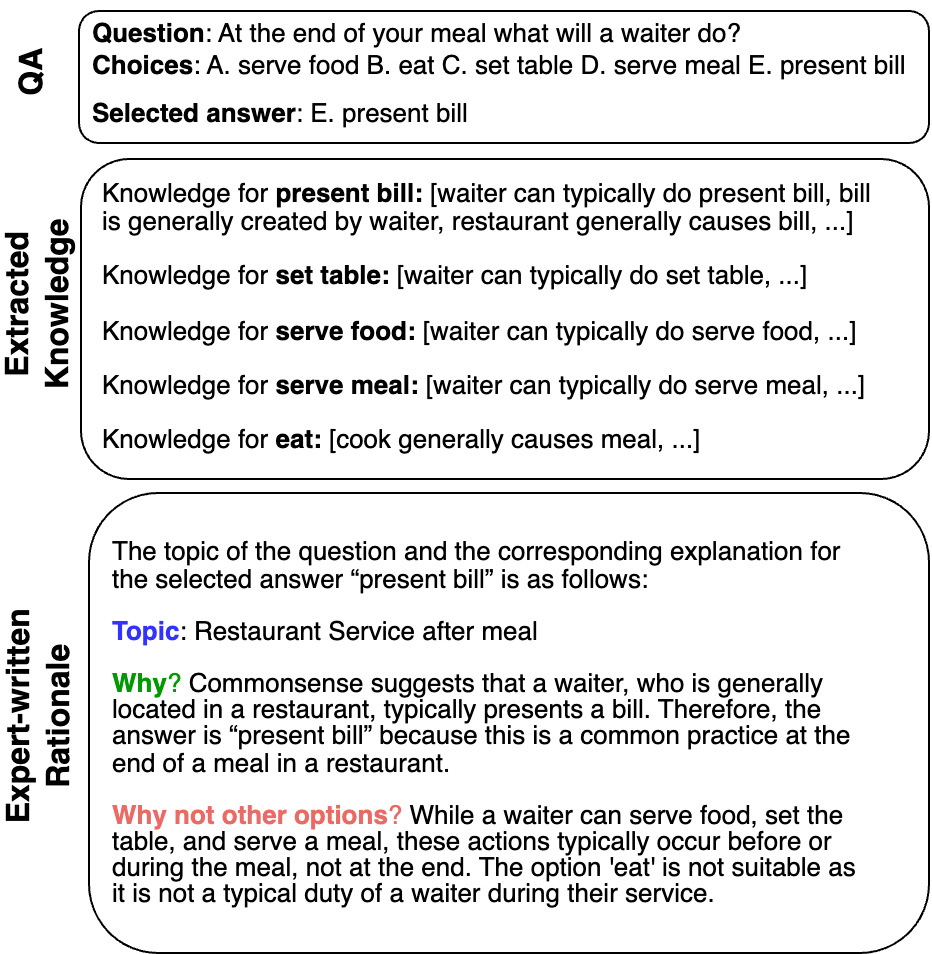}
    \caption{An example in the few-shot prompt: the QA and External Knowledge components are retrieved and the topic and the rationale are expert authored.}
    \label{fig:prompt}
\vspace{-10pt}
\end{figure}

\stitle{Prompt Design.} We employ greedy decoding-based few-shot prompting to query an LLM for rationalization. Figure~\ref{fig:prompt} outlines the few-shot prompt structure. Each example in the prompt can be divided into three components: input, knowledge, and demonstrations. The input refers to the original QA task and the KIT model's prediction. The knowledge component consists of the retrieved facts corresponding to each choice.
The demonstrations, written by humans (experts in this case), consist of a question topic followed by a rationale of the prediction grounded on the facts and consisting of corroboration and refutation. We opted for expert-authored rationales based on their reported effectiveness over crowdsourced rationales~\cite{wiegreffe-etal-2022-reframing}. The paper's authors collaboratively crafted high-quality rationales to compile the expert-written pool. Give an unseen multiple-choice question (\ie not included in the expert-written example pool); we combine the question, the model prediction, and the corresponding extracted facts with the sampled few-shot examples to formulate the final prompt (see Figure~\ref{fig:rationalizer}.)

\stitle{Rationale generation.}  We prompt the LLM to generate a topic of the question and a rationale similar to the provided few-shot examples. 
Therefore, our approach explicitly conditions the rationale generation on the question topic and the knowledge facts.
FARM~\cite{mei2023foveate} employs a similar topic-focused generation for question answering. Given a question, the LLM is initially prompted to generate a question context --- augmented with information retrieved from trustworthy sources --- to generate a safe response. 
Such strategies have been shown to be very effective~\cite{radford2019language, brown2020language, shin2020autoprompt, schick2020exploiting}, even in complex generation tasks~\cite{reif2021recipe}.

\section{Evaluation of Rationales}
\label{sec:study}
Due to a lack of suitable automated methods for evaluating the rationale quality~\cite{clinciu-etal-2021-study, kayser2021vil} and credibility, we conducted three studies to address the following questions:

\stitle{RQ1}. How preferable are the generated rationales to humans compared to crowdsourced datasets of knowledge-intensive tasks? ($\S$~\ref{sec:comparison})

\stitle{RQ2}. To what degree do the fine-grained properties of a rationale influence its acceptability? ($\S$~\ref{sec:acceptability})

\stitle{RQ3}. How does faithful rationalization of model predictions impact humans' trust in the LLM-generated rationales?  ($\S$~\ref{sec:trust})

\stitle{Datasets and Prompts.}
We select QAGNN~\cite{qagnn} as the KIT model due to its well-documented code repository and availability of pre-trained model weights. We consider two datasets of multiple-choice QA tasks related to commonsense knowledge, CSQA~\cite{csqa}, and elementary-level science, OBQA~\cite{obqa}. 
Following the existing KIT models, we use ConceptNet~\cite{speer2017conceptnet} as our external knowledge source. For both datasets, we report
results on a fixed, randomly-sampled 250-instance
test set. We sample these instances from the test set prepared for
these datasets~\cite{mhgrn}. We employed GPT-3.5 \code{text-davinci-003} ($temperature=0$) as the rationalizer. We randomly selected 40 instances from each of the CSQA and OBQA datasets --- different from the 250 test instances --- to be included in the example pool. See  Appendix~\ref{app:prompt} for details.  

\stitle{Faithful Rationalization Studies.}
We conducted two crowdsourced studies aimed at addressing \emph{RQ1} and \emph{RQ2}. For both studies, we only consider rationalization of correct KIT model predictions, \ie faithful rationalization. The approach is similar to prior work~\cite{aggarwaletal2021ecqa, wiegreffe-etal-2022-reframing, marasovic-etal-2022-shot,kayser2021vil} that also removed the confounder, \ie rationalization of incorrect model prediction,
by only considering rationales for correctly predicted instances. We used Amazon Mechanical Turk for crowdsourcing evaluation. For HITs in both studies, we asked targeted questions to obtain coarse- and fine-grained feedback on the rationales of a KIT model decision. 
We detail these evaluation metrics in the respective sections discussing the studies. Due to the subjectivity of some of the instances of the CSQA dataset, following Wiegreffe~\cite{wiegreffe-etal-2022-reframing}, we instruct workers for both the studies to consider the KIT model prediction to be correct even if they disagree with it. We also ensured that each experiment
had a substantial number of distinct crowd workers
to mitigate individual annotator bias.
We include the study interface design and other statistical information in Appendix~\ref{app:crowd} and Appendix~\ref{sec:screenshots}.


\stitle{Credible Rationalization Study.}
To address \emph{RQ3}, we conducted a preliminary study to confirm observations of existing work on trust in explainable AI~\cite{hoffman2018metrics, stites2021sage, smith2020no} in the context of LLM-generated rationales such as agreement, confidence, reliability, and user satisfaction, among others. In this study, we consider rationales generated on both correct and incorrect KIT model predictions. The study was conducted via a Slack campaign within \company, an industrial research lab, with natural language processing, data management, and machine learning as the primary research areas. 

\section{LLMs vs Humans as Rationalizers}
\label{sec:comparison}
We compare LLM-generated rationales of the  CSQA~\cite{csqa} tasks with corresponding crowdsourced rationales in ECQA~\cite{aggarwaletal2021ecqa} --- each ECQA rationale corroborate the selected answer in CSQA and refutes the other choice. We exclude CoS-E~\cite{rajani-etal-2019-explain}, another crowdsourced free-text rationales dataset, as those rationales are not refutation complete.

\subsection{Study Setting} 
\label{sec:comparison_metric}

In each of the 250 HITs (three different crowd-workers per HIT), a crowd-worker was presented with a question with choices, the corresponding prediction of the KIT model, and two rationales:  LLM-generated (from our pipeline) and human-written (from ECQA.) We then ask them to make a preferential selection among the two rationales (see interface details in Appendix~\ref{sec:screenshots_faith}.) We find
low-to-moderate annotator agreement -- Krippendorff's $\alpha =0.13$~\cite{krippendorff2011computing} --- for this study, indicating the subjective nature of the task. Related work~\cite{wiegreffe-etal-2022-reframing} reported similar agreement statistics ($\alpha \in [0.05, 0.20]$) on comparison between LLM-generated and ECQA rationales.  

\stitle{Fine-grained comparison.} Besides head-to-head comparison, we ask several 7-point Likert scale questions --- adapted from prior work~\cite{aggarwaletal2021ecqa, wiegreffe-etal-2022-reframing} --- targeted at comparing fine-grained aspects of both rationales. These aspects include: 
\emph{sufficiency} in justifying the model's choice;
\emph{conciseness} (\ie degree of redundancy);
\emph{understandability};
\emph{factuality} (\ie factual correctnes); 
\emph{supportiveness} (\ie the degree to which the model prediction is supported);
\emph{refutation convincingness} (\ie the degree to which the unselected choices are convincingly refuted);
\emph{insightfulness} (\ie how much new information is captured.) New information can be new facts or reasoning not stated in the question and answer choices and potentially grounded on the knowledge evidence. We report the agreement statistics on individual aspects in the Appendix~\ref{app:annotation_stat}.



\subsection{Higher Preference of LLM Generations}
Surprisingly, LLM-generated rationales were more frequently preferred ($67.2\%$ times) over human-written rationales ($37.8\%$ times.) 
The result showcases an improvement over previous work on generating corroboration only (no refutation) rationales~\cite{wiegreffe-etal-2022-reframing} --- $45.7\%$ preference to LLM generations.
The human-written ECQA rationales potentially outperformed those LLM-generated rationales on dimensions such as refutation convincingness, sufficiency, and supportiveness. Moreover, our pipeline enabled knowledge-guided rationale
generation, whereas prior LLM-generated rationales~\cite{wiegreffe-etal-2022-reframing} lacked such grounding and were abstractive. However, the LLMs in both studies differed, with our study employing a newer version (GPT-3.5) than the GPT-3 model used in their work.
While some of the performance gain
can be attributed to such model upgrades,
Next, we demonstrate via fine-grained analysis how aspects of our rationale construct are correlated with crowd-worker preference.


\begin{figure}[!htb]
    \centering
    \includegraphics[width=\linewidth]{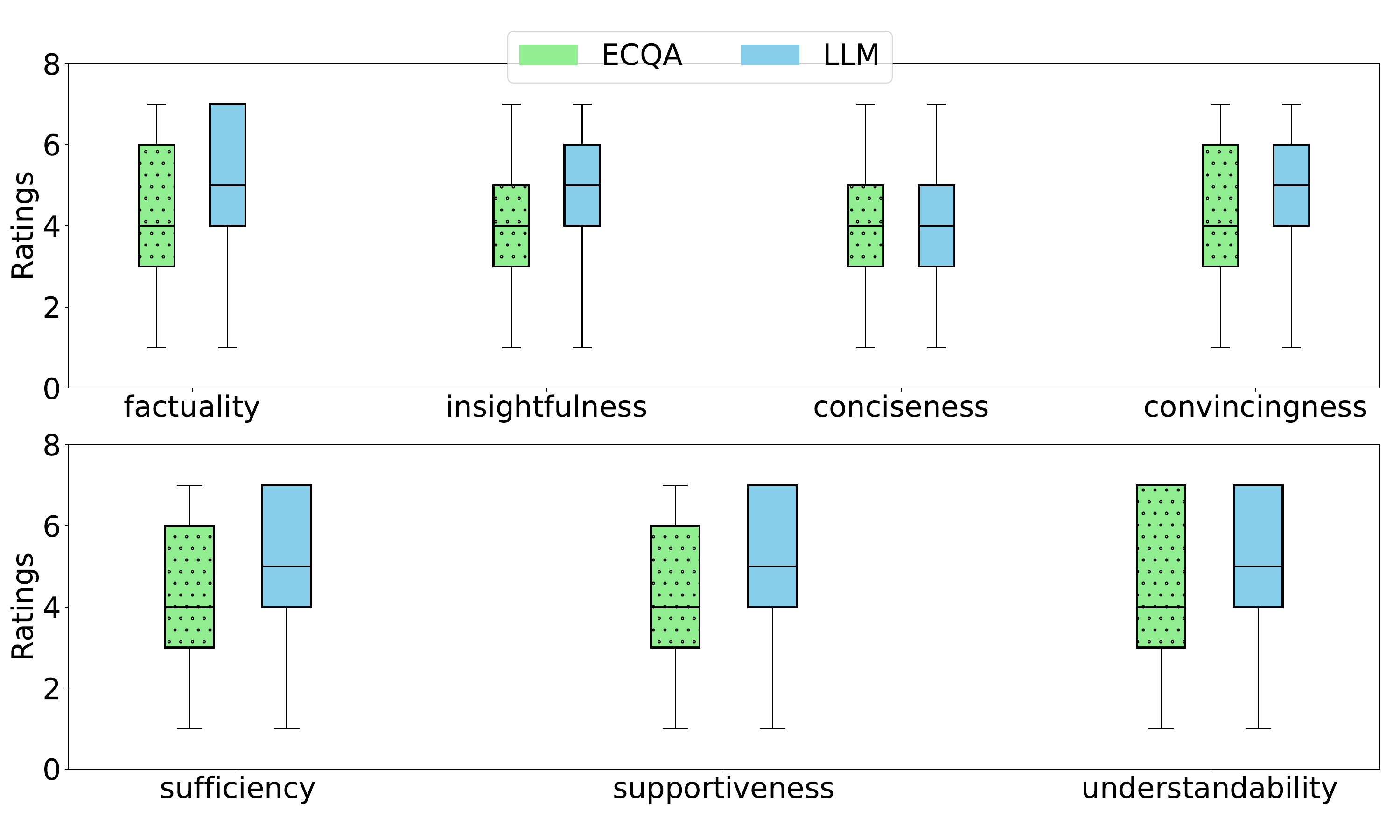}
    \caption{Distribution of fine-tuned metrics between human-written (ECQA) and LLM-generated rationales --- LLM-generated rationales were preferred over ECQA on majority of the metrics except conciseness.}
    \label{fig:rationalizer-comparison-fine}
\vspace{-10pt}
\end{figure}

\subsection{Fine-grained Comparison}
As shown in Figure~\ref{fig:rationalizer-comparison-fine}, overall, crowd workers exhibited more preference for LLM-generated rationales over human-written ones on aspects such as insightfulness (\ie new information), refutation convincingness, and sufficiency. In fact, up to $80.4\%$ of the LLM-generated rationales presented to the crowdworkers contained at least one statement grounded on external knowledge, thereby contributing to insightfulness (see Appendix~\ref{app:add_exp_guidance} for details.) Moreover, the refutation argument anchored on the topic of the question enabled a more convincing refutation. Therefore, the resulting LLM-generated rationales were sufficient to justify the model's choice for the QA task. 

The preference for LLM-generated and human-written rationales was comparable for other aspects such as factuality, supportiveness, and understandability. However, crowd-workers rated LLM-generated rationales as more redundant compared to human-written rationales. Such an observation is unsurprising, given the tendency of the LLMs to generate verbose text. While recent work has demonstrated the effectiveness of controlling such generation using carefully crafted prompts in zero-shot setting~\cite{liu2023gpteval}, such prompt optimization efforts through trial-and-error experiments are beyond the scope of this work.

\begin{table}[!htb]
\centering
\scriptsize
\begin{tabular}{c c c}
\hline
\textbf{Metrics}    & \textbf{LLM-generated } & \textbf{Human-written }         \\
 & \textbf{Preferred} & \textbf{Preferred} \\
\hline
Factuality              & 0.29                                & 0.04 \\
Insightfulness        & 0.21                                 & 0.12 \\
Conciseness             & 0.08                                 & 0.02 \\
Convincingness             & 0.29                                  & 0.17 \\
Sufficiency             & 0.28                                & 0.14 \\
Supportiveness                 & 0.27                                 & 0.03 \\
Understandability       & 0.27                                & 0.01 \\
\hline
\end{tabular}
\caption{Spearman correlation between crowdworker preference of rationale type. \ie LLM-generated and human-written. All the correlations are observed with $p < 0.01$ \textbf{indicating strong statistical significance}.}
\label{tab:comparative-correlation}
\end{table}

\stitle{Correlation to rationale preference.}
To understand what factors are important for the \emph{preference} judgment, we
compute Spearman correlation~\cite{spearman1987proof}
between the binary preference of both rationale types --- \ie LLM-generated and Human-written --- and the fine-grained aspects (see Table~\ref{tab:comparative-correlation}.) 
The conciseness aspect lacked any correlation with either rationale type. Surprisingly, crowd-workers' preference for human-written rationales lacked any correlation with several other aspects, such as factuality, supportiveness, and understandability, while showcasing a very weak correlation with the rest.  
However, these fine-grained aspects exhibited a stronger positive correlation with the LLM-generated rationales. 
Further analysis showcases that even when crowd-workers preferred ECQA rationales in a head-to-head comparison, LLM-generated and human-written rationales exhibited almost similar ratings in the majority of the fine-grained aspects (see Appendix~\ref{app:llm_ecqa}.)
Overall, the results indicate that human preference for LLM-generated rationale can be captured by factoring in different fine-grained aspects, which can inform the design of automated mechanisms for estimating the suitability of a rationale for end-users. 



\section{Acceptability of LLM Rationalization}
\label{sec:acceptability}
While pairwise evaluations of preferences provided 
perspective on the relative quality of the 
rationales, we conducted another study 
to independently measure the acceptability of the LLM-generated rationales and collect absolute crowd-worker judgments across several aspects
related to rationale quality for both CSQA and OBQA datasets.

\subsection{Study Setting}
In each of the 250 HITs per dataset (three different judges per HIT), a crowd-worker was presented with a question with choices, the corresponding prediction of the KIT model, and an LLM-generated rationale. Besides asking 7-point Likert scale questions on fine-grained aspects of a rationale --- similar to the first study in Section~\ref{sec:comparison} --- we include two additional surface-level aspects: \emph{readability}, \ie the clarity of the provided justifications and \emph{grammaticality}, adherence to grammatical rules. Finally, we ask for an overall judgment on quality, \ie the overall acceptability of a rationale (see interface details in Appendix~\ref{sec:screenshots_faith}.) We again find low-to-moderate agreement -- Krippendorff's $\alpha = 0.12$ for CSQA and $0.15$ for OBQA dataset. 
Related work~\cite{wiegreffe-etal-2022-reframing} reported slightly better agreement statistics ($\alpha =0.28$) on the CSQA dataset (see Appendix~\ref{app:annotation_stat} for details.)

\subsection{Favorability Towards LLM generations}
On the overall acceptability metric, the LLM-generated rationales received a notably positive rating from the participants for both CSQA ($\mu=5.83, \sigma=1.27$) and OBQA ($\mu=5.89, \sigma=1.50$). These independent observations reaffirm earlier takeaways ($\S$~\ref{sec:comparison}) and underscore that the LLM-generated rationales of KIT models were viewed favorably by crowd-workers.



\begin{figure}[t]
    \centering
    \includegraphics[width=\linewidth]{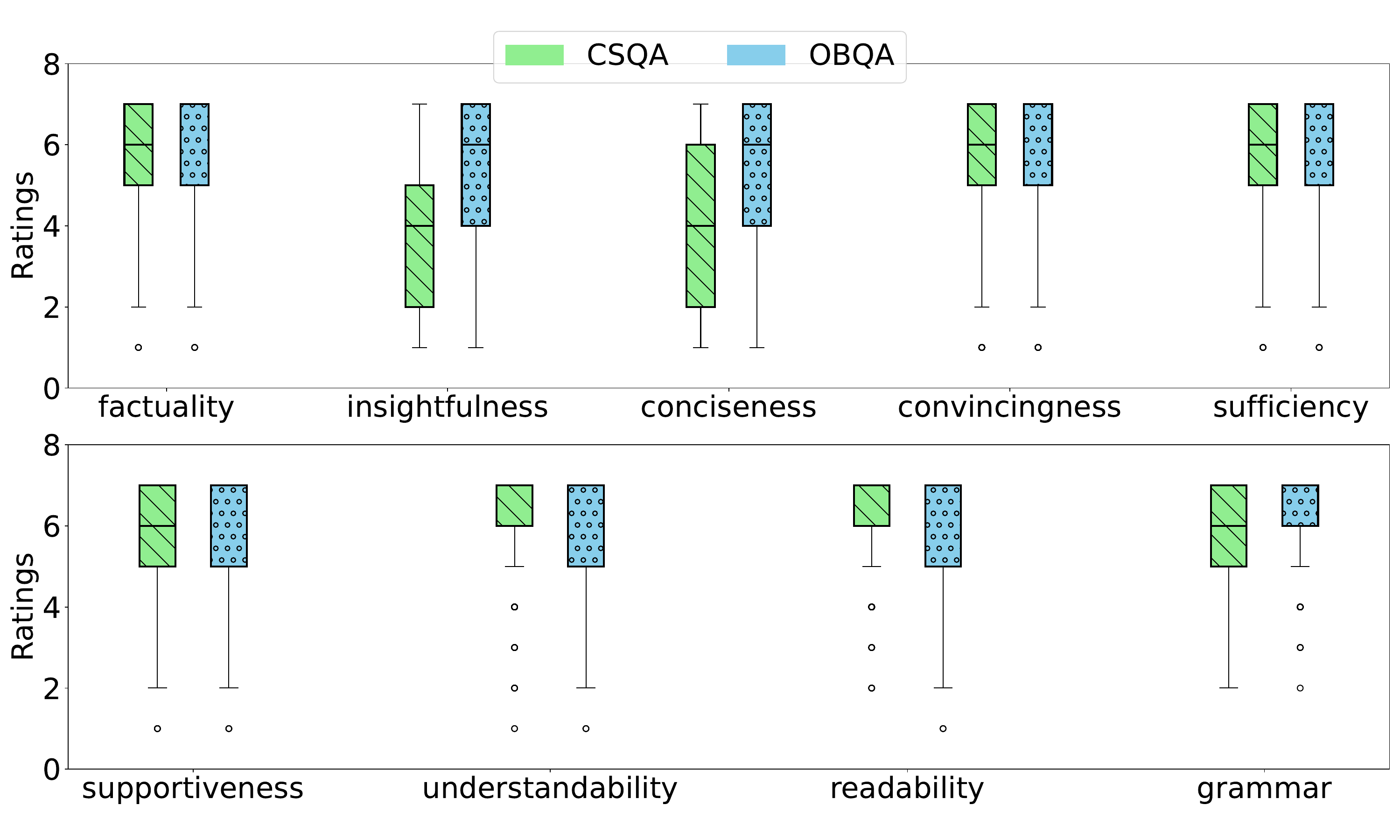}
    \caption{Crowdworkers' ratings showed similar distrbution for all metrics except insightfulness and concisenes. These metrics were rated lower for the more subjective CSQA dataset compared to the objective and scientific OBQA dataset.}
    \label{fig:rationalizer-acceptable-csqa}
\vspace{-10pt}
\end{figure}

\stitle{Fine-grained observations.} 
As shown in Figure~\ref{fig:rationalizer-acceptable-csqa}, 
for the newly introduced surface-level metric, readability, and grammaticality, the LLM-generated rationales received higher ratings in keeping with the previous work. In fact, for both datasets, for all of the richer aspects except \emph{insightfulness} and \emph{conciseness}, the ratings received were similar, \ie more positively rated. 
While the insightfulness metric was rated positively for OBQA, the rating was neutral to slightly negative for CSQA.
Surprisingly, conciseness (\ie less redundancy) was rated positively for OBQA, whereas CSQA rationales were deemed more redundant, similar to the previous study.
A plausible explanation for this discrepancy is the inherent subjectivity in CSQA~\cite{wiegreffe-etal-2022-reframing}, which can result in varying expectations regarding the information provided in the rationales. In contrast, the OBQA dataset is grounded in objective scientific facts, eliminating such subjectivity and leading to more consistent expectations among crowd-workers.

\begin{table}[!htb]
\centering
\scriptsize
\begin{tabular}{ccc}
\hline
\textbf{Metrics}    & \textbf{Correlation CSQA} & \textbf{Correlation OBQA}         \\
\hline
Factuality              & 0.65                                & 0.73 \\
Insightfulness        & 0.38                                 & 0.67 \\
Conciseness              & 0.09                                 & 0.6 \\
Convincingness & 0.70                                  & 0.80 \\
Sufficiency             & 0.76                                 & 0.80 \\
Supportiveness                & 0.54                                 & 0.76 \\
Understandability       & 0.63                                 & 0.71 \\
Readability             & 0.5                                  & 0.74 \\
Grammar                 & 0.33                                 & 0.62
\\
\hline
\end{tabular}
\caption{Spearman correlation between acceptability and the fine-grained aspects of a rationale were observed with \textbf{strong statistical significance} ( $p < 0.01$).}
\label{tab:acceptable-correlation}
\end{table}

\stitle{Correlation to overall acceptability.} 
To understand what factors are important for the overall \emph{acceptability} judgement, we
compute Spearman correlation~\cite{spearman1987proof}
between \textit{acceptability} and the fine-grained aspects (see Table~\ref{tab:acceptable-correlation}.) For both the 
datasets, all aspects except \emph{conciseness} show similar patterns --- varying degrees of positive correlation with acceptability. However, the rationales for the CSQA dataset (more subjective) exhibited a weaker correlation than the OBQA dataset rationales (more objective) in several aspects, such as  
conciseness, insightfulness, readability, and grammaticality.
The results indicate that human preference for rationale is more nuanced and can only be holistically captured by considering different fine-grained aspects. However,
the quality of the generated rationale may vary depending on the task and domain type and, consequently, impact human-preference judgment.
Therefore, there is further room for improvement in making generated rationales invariant to task and domain.


\section{Towards Credible Rationalization}
\label{sec:trust}
In the earlier studies, similar to existing work~\cite{aggarwaletal2021ecqa, wiegreffe-etal-2022-reframing, marasovic-etal-2022-shot,kayser2021vil}, we evaluate LLM-generated rationales for cases where model prediction matches the ground truth. We now investigate the implications of rationalization without accounting for model errors, \ie faithful rationalization, and potential intervention strategies. 

\subsection{Trustworthiness of Generated Rationales}
The reported accuracy of KIT models widely vary --- from $64\%$-$89.4\%$ for the CSQA datasets\footnote{\url{https://www.tau-nlp.sites.tau.ac.il/}} and $64\%$-$89.4\%$ for the OBQA dataset~\footnote{\url{https://leaderboard.allenai.org/open_book_qa/}}. The reported human accuracy for the CSQA and OBQA datasets are $88.9\%$ and $91.7\%$, respectively. Even as humans rationalize, the credibility of the rationalizer may diminish if they attempt to justify any incorrect decisions. Existing work on trust in explainable AI (XAI) literature~\cite{hoff2015trust, schaefer2016meta, stites2021sage, smith2020no} demonstrates that end-users' familiarity and prior experience in a domain often impact their trust --- trust in a system degrades when encountering errors they can easily recognize.
We also aim to investigate the relationship between model accuracy and humans'
degree of trust in the context of free-text rationales, an unexplored aspect in explainable NLP literature. Since the knowledge source for the CSQA and OBQA datasets is ConceptNet~\cite{speer2017conceptnet}, a commonsense knowledge graph, humans are expected to have higher confidence about their knowledge in the domain. We replicate the study design of exploring trust in explanations for classification models~\cite{stites2021sage} to confirm whether the observations hold for knowledge-intensive QA tasks in the commonsense domain.

\stitle{Study design.}
We conducted a between-subject study involving $22$ participants
($15$ male and $7$ female) exploring two conditions: $66\%$ ($11$ participants) and $90\%$ ($11$ participants) model accuracy. The accuracy conditions reflect the two extremities of existing knowledge-intensive task models~\cite{qagnn, mhgrn, dragon}.
The study consisted of three phases: an introduction to the study, a quiz phase, and a follow-up survey. In the quiz phase, the participants answered 15 QA tasks. The 15 tasks were randomly selected from the CSQA (8 QAs) and OBQA (7 QAs) datasets. Depending on the study conditions, for $X\%$ of those $N$ questions, where $X \in \{66, 90\}$, the KIT model made accurate predictions, and the rest of the predictions were inaccurate.
The KIT model prediction and LLM-generated rationale of a QA task were revealed \emph{after} a participant submitted their response to avoid bias. Then, the participants were asked whether they agreed with the model prediction and had to rate their impression of the rationale on a scale of 1 to 7 (1 = actively misleading and 7 = helpful.) After the quiz phase, the participants completed a survey adapted from the Trust Scale recommended for XAI~\cite{hoffman2018metrics}. The survey contained questions that asked participants to rate several aspects related to the quiz phase tasks, such as the goodness of rationales, satisfaction with the rationales, and the participants' trust and reliance on the LLM-generated rationale. All of these required participants to work slowly enough to be able to read all the items, thereby making the studies long-running and rather unsuitable for crowd platforms according to existing work~\cite{douglas2023data}. Therefore, we opted for internal recruitment as an additional quality control mechanism, inviting participants internally via a Slack campaign at \company. None of the participants are authors of the paper. We discuss the study design in more detail in Appendix~\ref{app:trust_study}.

\subsection{Preliminary Study Results} 

\begin{figure} 
\vspace{-10pt}
    \centering
  \subfloat[\label{xai_rationale_a}Impact on user perception]{%
       \includegraphics[width=0.5\linewidth]{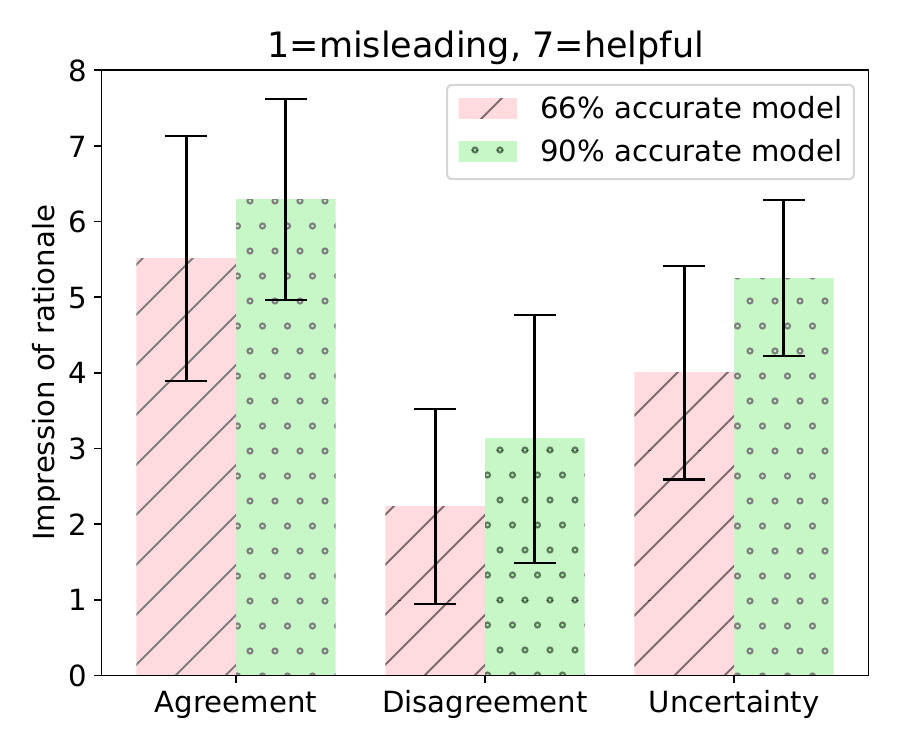}}
    \hfill
  \subfloat[\label{xai_rationale_b}XAI Trust Scale feedback]{%
        \includegraphics[width=0.5\linewidth]{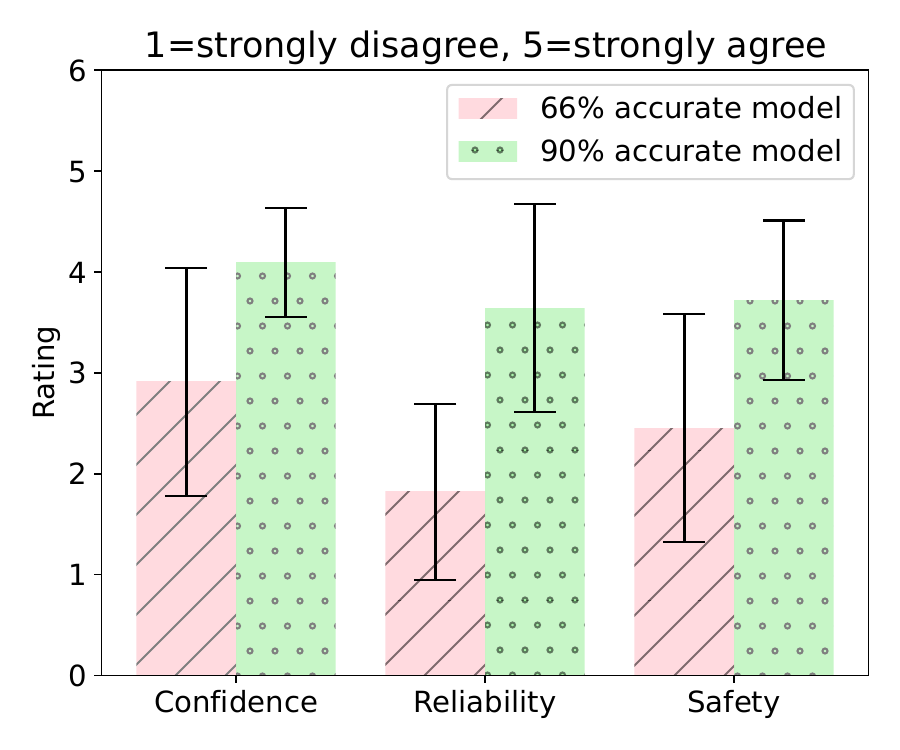}}
  \caption{(a) Irrespective of agreement or disagreement with the KIT model prediction, participants indicated a more negative impression about the rationalization of the lower confidence model prediction. (b) Participant feedback on trust scale indicates lower confidence for lower accuracy model rationalization.}
  \label{fig:xai_rationale} 
\vspace{-10pt}
\end{figure}

The agreement statistics of the participants reflect both the study conditions --- $67.27\%$ and $86.07\%$ for lower and higher accuracy models, respectively. 
Figure~\ref{fig:xai_rationale}a summarizes the participants' impression of a rationale immediately after viewing the model prediction. When the participants disagreed with the model prediction, they exhibited a stronger negative impression about the rationales for the $66\%$ accuracy condition compared to the $90\%$ accuracy condition. Even when participants agreed with the model prediction, their impression of the rationales remained more negative. Our intuition is that the higher disagreement with the model coupled with observing the faithful rationalization of the incorrect prediction negatively impacted participants' perception of the reliability of the rationales. We confirm these observations by analyzing the results of the follow-up survey (see Figure~\ref{fig:xai_rationale}b.)  Unsurprisingly, participants for the $66\%$ accuracy condition rated their confidence in the generated rationales and the reliability of the rationalizer significantly lower compared to the $90\%$ accuracy condition. The trends in Figure~\ref{fig:xai_rationale} are observed with strong statistical significance, except for participant feedback on satisfaction with rationale (see Appendix ~\ref{app:trust_exp_additional}.)

\subsection{A Review-then-Rationalize Framework}
\begin{figure}[!htb]
    \centering
    \includegraphics[width=\linewidth]{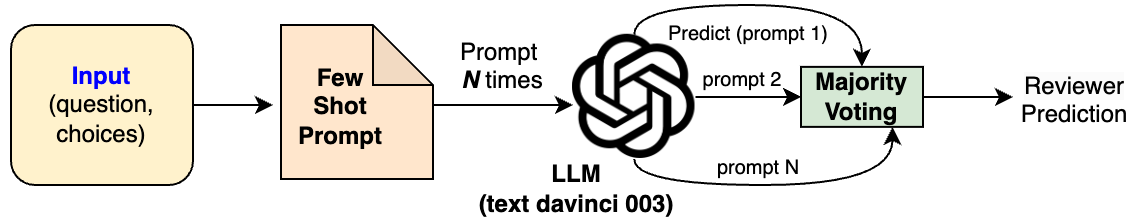}
    \caption{Self-consistency-based Reviewer---intervene for any disagreement with the KIT model prediction.}
    \label{fig:acc-rationalizer}
\end{figure}

Motivated by the observations from the preliminary study,
we create a two-stage review-then-rationalize (see Figure~\ref{fig:acc-rationalizer}) pipeline to intervene 
incorrect model predictions before rationalization. 
The pipeline instruments a \emph{reviewer} module
that employs another model (GPT-3.5 \code{text-davinci-003} (\code{temperature = 0})) to evaluate the correctness
of the knowledge-intensive task model and refrain from
rationalizing potentially incorrect decisions. 
 
 We opted for LLMs as reviewers due to their reported proficiency in natural language understanding. Depending on the task and data domain, the suitability of the reviewer model may vary. Given the complexity of knowledge-intensive tasks, 
 we employ a self-consistency-based decoding strategy~\cite{wang2022self}
 as opposed to greedy-decoding to ensure robustness. More specifically, we independently pose the same QA task $N$ (=5) times to the reviewer and select the final response via majority voting. The reviewer then compares the model's prediction with its prediction and activates the rationalizer only when both models agree. A cookie-cutter rationale or no rationale may be communicated to the end-user in a disagreement.

\begin{table}[!htb]
\scriptsize
\begin{tabular}{cccc}
\hline
\multicolumn{1}{c}{\multirow{2}{*}{\textbf{Dataset}}} & \multicolumn{1}{c}{\textbf{Prediction Errors}} & \multicolumn{2}{c}{\textbf{Errors Intervened}}                                                \\ \cline{3-4} 
\multicolumn{1}{c}{}                                  & \multicolumn{1}{c}{\textbf{(Test Set)}}        & \multicolumn{1}{c}{\textbf{Greedy Decoding}} & \multicolumn{1}{c}{\textbf{Self-consistency}} \\ \hline
CSQA                                                    & 321                                             & \multicolumn{1}{c}{166 ($51.71\%$)}                     & 187 ($\mathbf{58.26\%}$)                                        \\ 
OBQA                                                    & 155                                             & \multicolumn{1}{c}{102 ($65.81\%$)}                     & 110 ($\mathbf{70.97\%}$)                                         \\ \hline
\end{tabular}
\caption{The review-then-rationalize pipeline helps intervene incorrect predictions of a knowledge-intensive task model. The self-consistency-based reviewer outperforms the greedy decoding-based reviewer.}
\label{tab:review}
\end{table}

As shown in Table~\ref{tab:review}, for knowledge-intensive tasks such as CSQA and OBQA, the proposed pipeline helps intervene up to $58\%$ and $71\%$ of the incorrect predictions. Unsurprisingly, the self-consistency-based reviewer outperforms the greedy decoding-based reviewer. The results draw attention to the fact that such intervention is crucial when rationalizing a model's decision to end-users. We focused on developing a general \textit{intervention} strategy for the setting and showcasing its effectiveness in accounting for incorrect model predictions. Future work may explore different communication strategies during prediction errors, such as communicating the disagreement to the experts-in-the-loop,  providing rationales with a disclaimer, and employing stronger reviewers to repair the prediction on the fly and then rationalize, among others.


\section{Related Work}
\label{sec:related}

\stitle{Free-text Rationale Generation.}
Existing works highlight the effectiveness of free-text rationales in justifying a model's decision to humans in vision~\cite{hendricks2016generating, park2018multimodal} and text domains~\cite{camburu2018snli, ehsan2018rationalization, narang2020wt5}. 
Due to cost and generalizability implications
of supervised rationale generation,
we employ few-shot prompting to elicit rationales 
from LLMs following existing work~\cite{wiegreffe-etal-2022-reframing, marasovic2021few}. 
Both these approaches generate abstractive, corroborative, and faithful rationales.
In contrast,
we explore the generation of knowledge-guided, corroborative and refutation complete, and credible 
rationales.

\stitle{Guided text generation.}
Developing approaches to avoid hallucinations and factual inaccuracies in LLM-generated text is a new area of research. Retrieval augmented generation (RAG) infuses 
external knowledge ~\cite{peng2023check, lazaridou2022internet}, such as knowledge-bases and web documents, while prompting LLMs to help generate responses
. We employ a similar strategy during rationalization by conditioning the LLM generation on the retrieved evidence for a given task.

\stitle{Credible text generation.} Studies in explainable AI literature~\cite{smith2020no, hoff2015trust,schaefer2016meta,stites2021sage} demonstrate that for low-quality models, providing faithful explanations further degraded user's trust. 
Unlike existing work on free-text explanation~\cite{wiegreffe-etal-2022-reframing, marasovic2021few}, we explore how end-users' trust may be impacted by faithful rationalization of varying degrees of incorrect model predictions. ReXC~\cite{majumder2021knowledge} augments rationales --- generated in a self-rationalization framework --- with background knowledge to improve a model's task performance, such as natural language inference and visual commonsense reasoning. 
To rectify incorrect LLM responses, identified via a self-consistency-based intervention approach, the Verify-then-Edit framework~\cite{zhao2023verify} leverages external knowledge to repair reasoning chains of the corresponding chain-of-thought prompts. FARM~\cite{mei2023foveate} utilizes trustworthy 
external sources within a predict-then-generate framework that aims to intervene in harmful content generation using LLMs. To credibly rationalize KIT model predictions, we explore a review-then-rationalize framework where
a self-consistency-based reviewing approach identifies potential prediction inaccuracies and ensures credible rationale generation.

\section{Conclusion}
\label{sec:conclusion}
We demonstrate an LLM’s capacity 
to generate effective rationales for 
knowledge-intensive tasks in a few-shot knowledge-guided setting.
We additionally investigate the implications of employing
the LLM as a rationalizer of an imperfect model and highlight the negative impact on users' trust.
Observations from our studies highlight room for improvement
in aspects such as task and domain invariant rationalization and robust intervention 
strategies for real-world usage.

\section{Limitations}
\label{sec:limitations}

\stitle{Scrutinizing LLM-generated rationales.} 
While external knowledge-guided generation offers promise~\cite{peng2023check, mallen-etal-2023-trust}, LLM-generated rationales may still suffer from 
hallucinations. Our experiments highlight that the LLM-generated rationale is not entirely grounded on retrieved knowledge. Even though crowd-workers positively rated the factuality and insightfulness of the generated rationales, additional scrutiny is required before deploying such rationalizers in mission-critical tasks. To this end, 
the review-then-rationalize framework may be expanded to further scrutinize the rationales by 
adopting recent work on an LLM's factual knowledge measurement~\cite{pezeshkpour2023measuring, dong2023statistical} and hallucination identification~\cite{manakul2023selfcheckgpt,elaraby2023halo,mundler2023self} and reduction~\cite{zhao2023verify,mei2023foveate}, and explainable evaluation~\cite{xu2023instructscore}.

\stitle{Fairwashing vs. credible rationalization.}
The accuracy of our self-consistency-based reviewer can be further improved
to intervene in a higher proportion of incorrect KIT model predictions.
However, critiques of XAI tools have raised concerns about
\emph{fairwashing}, \ie misleading users
into trusting biased or incorrect models~\cite{alikhademi2021can}.
For example, simply averting potential
faithful yet incorrect rationalization, identified by the reviewer, may increase end-users' trust
due to an illusion of a highly performant rationalizer~\cite{aivodji2019fairwashing}.
Such fairwashing may have catastrophic consequences
if employed in real-world applications such as 
in the medical domain, hiring platforms, and credit agencies.
Recent work~\cite{alikhademi2021can}
proposes a framework for evaluating XAI tools with 
respect to their capabilities for detecting and addressing issues of bias and fairness
as well as their capacity to communicate these results to their users
clearly. Therefore, future implementations of the credible 
rationale should adopt similar strategies to safeguard against
such issue. 




\stitle{Scaling responsibly.}
An often overlooked aspect of the recent popularity of LLMs has been \emph{Green AI}~\cite{schwartz2020green}. While the environmental impact and financial cost of pre-training of such large language models have been highlighted by existing work~\cite{10.1145/3442188.3445922}, employing LLMs to scale up downstream applications such as rationale generation can have similar effects. As open-source LLMs continue to improve, closed-source LLMs such as GPT-3~\cite{brown2020language} remain more popular. Even in designing our proposed GPT-3.5-based rationalizer, we considered budget (GPT-4 is costlier than GPT-3.5) and efficiency (cost of running open-source LLMs on-premise compared to their lower accuracy.) Future studies may explore these dimensions across models of varying performance and size. Moreover, when the ML deployment pipeline is considered as a whole, inference consumes most compute resources, accounting for anything between 70$\%$ to 90$\%$~\cite{weng2022mlaas,wu2022sustainable}. Knowledge distillation approaches can be adopted to avoid costly pre-training~\cite{wang2023scott}. Furthermore, materialization
of rationales to avoid repeating rationalizing the same task can be possible approaches to handle such issues.

\bibliography{reference}

\appendix

\section{Prompts and Rationales}
\label{app:prompt}

In this section, we provide additional details regarding the prompts corresponding to the faithful and Figure~\ref{fig:prompt}) and credible rationalization workflows. 

\begin{table*}[!htb]
\scriptsize
    \centering
\begin{tabular}{p{5.6in}}
\hline
Question: At the end of your meal what will a waiter do?\\
Choices: A. serve food B. eat C. set table D. serve meal E. present bill\\
Selected answer: E. present bill\\
\\
Knowledge for present bill: [waiter can typically do present bill, bill is generally created by waiter, restaurant generally causes bill, $\ldots$]\\
Knowledge for set table: [waiter can typically do set table, $\ldots$]\\
Knowledge for serve food: [waiter can typically do serve food, $\ldots$]\\
Knowledge for serve meal: [waiter can typically do serve meal, $\ldots$]\\
Knowledge for eat: [cook generally causes meal, $\ldots$]\\
\\
The topic of the question and the corresponding explanation for the selected answer “present bill” are as follows: \\
Topic: Restaurant Service after meal\\
Why? Commonsense suggests that a waiter, who is generally located in a restaurant, typically presents a bill. Therefore, the answer is “present bill” because this is a common practice at the end of a meal in a restaurant.\\
Why not other options? While a waiter can serve food, set the table, and serve a meal, these actions typically occur before or during the meal, not at the end. The option 'eat' is not suitable as it is not a typical duty of a waiter during their service.\\
$=======================================================================$\\
Question: He waited for his friend at the squash court, but he was worried his friend thought he meant the at the other end of the public what?\\
Choices: A. country club B. rich person's house C. pool D. park E. fitness center
Selected Answer: D. park\\
\\
Knowledge for park :[squash court is generally located in park, play is generally located in squash court, $\ldots$]\\
Knowledge for fitness center :[squash court is generally located in fitness center, $\ldots$]\\
Knowledge for country club :[squash court is generally located in country club, $\ldots$]\\
Knowledge for pool :[$\ldots$]\\
Knowledge for rich person's house :[$\ldots$]\\
\\
The topic of the question and the corresponding explanation for the selected answer “park” are as follows:\\ 
topic: Public Spaces and Miscommunication\\
Why? The answer is park because commonsense suggests that a squash court is generally located in a park. This implies that there could be another squash court at the other end of the park, leading to the friend's confusion.\\
Why not other options? While a squash court can also be located in a fitness center or country club, these locations are not typically public spaces with multiple ends. A pool or a rich person's house are less likely to have multiple squash courts, making them less likely to be the source of the friend's confusion.
\\
$======================================================================$\\
Question: What should the bean bag chair sit on?	\\
Choices: A. house B. den C. family room D. wood E. floor\\
Selected Answer: E. floor\\
\\
Knowledge for present floor: [$\ldots$]\\
Knowledge for house: [$\ldots$]\\
Knowledge for  den: [$\ldots$]\\
Knowledge for family room: [$\ldots$]\\
Knowledge for wood: [$\ldots$]\\
\\
The topic of the question and the corresponding explanation for the selected answer “present bill” are as follows: \\
\\ \hline
\end{tabular}
\caption{Example of a prompt with two training examples for CSQA and an unseen question for which the LLM generated a rationale. In practice, we provide five examples.}
    \label{tab:prompt_faith_example}
\end{table*}

\subsection{Faithful Rationalization}
Table~\ref{tab:prompt_faith_example} elaborates on the prompt design shown in (Figure~\ref{fig:rationalizer} and Figure~\ref{fig:prompt}). Each example in the few shot prompt includes the question
and answer choices, the KIT model selected answer, the knowledge facts extracted from ConceptNet for each choice, and the expert-written question topic and rationale that act as input to GPT-3.5 \code{text davinci 003}. While we show only two few-shot examples, in practice, we use five examples per prompt. As explained in the Section~\ref{sec:pipeline}, due the token limit imposed by the GPT-3.5 API, we can include from 5-8 examples depending on the length of the knowledge facts. Given the prompt, \ie examples followed by an unseen question and answer choices, KIT model selected answer, and extracted knowledge, the LLM greedily generates the question topic and the rationale for the model prediction.

To design the initial prompt, we take inspiration from existing work~\cite{wiegreffe-etal-2022-reframing, peng2023check, lazaridou2022internet, zhao2023verify} to experiment with the prompt layout. We experimented with approximately 6 different layouts in the OpenAI playground~\footnote{https://platform.openai.com/playground} using 10 training examples on the CSQA
and OBQA datasets. In deciding the number of few-shot examples, we considered the maximum context window size of GPT-3.5 \code{text-davinci-003}, which is 4097 tokens. We observed that depending on the datasets and the length of the factual statements retrieved from ConceptNet, five to eight few-shot examples fit into the token constraints.
After finalizing the prompt layout, we developed a pool of $40$ expert-written (\ie authors of these papers) examples. We randomly selected $5$ expert-written examples for each test instance to ensure uniformity across datasets and instances.
Similar to prior work~\cite{wiegreffe-etal-2022-reframing}, we focused on developing a general few-shot prompting strategy for generating knowledge-enhanced and refutation complete rationale that could be successful when no additional (large) validation set for parameter tuning is available.

\subsection{Credible Rationalization}
Table~\ref{tab:prompt_cred_example} showcases the prompt design for the \emph{Reviewer} model within the credible rationalizer pipeline (Figure~\ref{fig:acc-rationalizer}. Each of the five examples in the few shot prompt includes the question
and answer choices that act as input to GPT-3.5 \code{text davinci 003}. In practice, we use five examples per prompt. Given the prompt, \ie examples followed by an unseen question and answer choices, the LLM greedily generates a response, \ie predicts an answer from the choices. We repeat the process five times and select a response based on majority voting. We randomly sample five questions from the 40 expert-written rationale pool as few-shot examples.

\begin{table}[!htb]
\scriptsize
    \centering
\begin{tabular}{p{2.8in}}
\hline
Question: At the end of your meal what will a waiter do?\\
Choices: A. serve food B. eat C. set table D. serve meal E. present bill\\
Selected answer: E. present bill\\
$==================================$\\
Question: He waited for his friend at the squash court, but he was worried his friend thought he meant the at the other end of the public what?\\
Choices: A. country club B. rich person's house C. pool D. park E. fitness center
Selected Answer: D. park\\
$==================================$\\
Question: What should the bean bag chair sit on?	\\
Choices: A. house B. den C. family room D. wood E. floor\\
Selected Answer:\\
\\ \hline
\end{tabular}
\caption{A prompt with two training examples for CSQA and an unseen question for the Reviewer to answer.}
    \label{tab:prompt_cred_example}
\end{table}

\subsection{LLM-generated Rationales}
Table~\ref{tab:rationale_faith_example} a few non-cherry picked examples of LLM-generated rationales. We show examples of rationales for CSQA dataset questions generated by both LLM and humans, \ie crowdworkers in the ECQA dataset. Since there is no crowdsourced dataset of OBQA rationales, we only show LLM-generated rationales. Note that the LLM greedily generates a topic of the question and a rationale with corroboration (``Why?'') and refutation (``Why Not'') components. However, as shown in Figure~\ref{fig:rationalizer}, the these two components are extracted from the generated output to construct the eventual rationales (\ie the formatting step at the end.)

\begin{table}[!htb]
\scriptsize
    \centering
\begin{tabular}{p{2.8in}}
$==================================$\\
CSQA Dataset\\
$==================================$\\
\textbf{Question}: What should the bean bag chair sit on?	\\
\textbf{Choices}: A. house B. den C. family room D. wood E. floor\\
\textbf{Selected Answer}: E. floor\\ \hline \\
\textbf{LLM-generated rationale:} The answer is floor because the commonsense knowledge clearly indicates that a bean bag chair is generally located in a floor.

While a bean bag chair can be placed in a house, den, family room, or on wood, the floor is the most common place for a bean bag chair to be located. \\ \\

\textbf{ECQA rationale:}  Bean bag chair is a seat people sit on which is generally put on a floor. A bean bag chair should sit on a floor and not on anything else from the other options.\\
$==================================$\\
OBQA Dataset\\
$==================================$\\
\textbf{Question}: Rainbows are always found after what?	\\
\textbf{Choices}: A. A fire B. A tornado C. Rainfall D. Cereal	\\
\textbf{Selected Answer}: C. Rainfall \\ \hline\\

\textbf{LLM-generated rationale:} The answer is Rainfall because rainbows are always found after rain. This is because the sunlight is refracted by the raindrops in the air, creating the rainbow. A fire, a tornado, and cereal do not have any relation to rainbows. \\ \hline
\end{tabular}
\caption{Rationales for CSQA generated by LLM and humans (ECQA) and for OBQA generated by LLM.}
    \label{tab:rationale_faith_example}
\end{table}

\section{Crowd Study Details}
\label{app:crowd}
We provide more details regarding both crowdworker studies
such as additional statistics related to the crowd study and quality control mechanisms. 

\subsection{Quality Control and Payment}
In order to enforce quality throughout evaluation,
we  use a hidden built-in Javascript
function to compute time per HIT spent and perform attention
checks by inserting questions with specific instructions randomly within 
a HIT. We filter out any annotator who completed the tasks in an unreasonably low
time, or failed their attention checks. To mitigate individual annotator bias, we also ensure that each experiment in a study has a substantial
number of distinct crowdworkers. See Tables~\ref{tab:compare_all_irr} and~\ref{tab:accpetanle_all_irr} for details regarding the inter-annotaror agreement for the comparison study.
For both studies, we used a pay rate of \code{USD} $12.00$/hr. We performed periodic check to ensure that the median HIT completion time remains commensurate to approximately the pay rate.
Median times reported for the comparative study was 208 seconds (paid at 80 cents each) the acceptability study was 110 seconds (paid at 40 cents each.) To ensure the quality of responses, we require annotators in Australia, New Zealand, United Kingdom, United States, and Canada as a proxy for English competency. We only selected workers with a past approval rate $>$ 98\% and who have completed over $5000$ HITs. We documented a worker's HIT submission time and performed attention checks within each HIT to enforce quality control. Note that each crowd worker was presented with detailed instructions about the study interface and performed an example task as a warm-up.

\begin{table}[t]
\scriptsize
    \centering
    \begin{tabular}{lcc}
    \hline
        \textbf{Approach} & \textbf{LLM-generated} & \textbf{ECQA} \\ \hline
        {Factuality} & 0.07 & 0.05 \\ 
        {Insightfulness} & 0.15 & 0.03 \\ 
        {Conciseness} & -0.04 & -0.01 \\ 
        {Convincingness} & 0.09 & 0.03 \\ 
        {Sufficiency} & 0.08 & 0.07 \\ 
        {Support} & 0.08 & -0.01 \\ 
        {Understandability} & 0.09 & 0.06 \\ 
        {Preference} & 0.13 & 0.13 \\ \hline
    \end{tabular}
\caption{Inter annotator agreement (Krippendorff's $\alpha$) of crowdworkers on the fine-grained aspects of a rationale evaluated in the head-to-head comparison study.}
    \label{tab:compare_all_irr}
\end{table}

\begin{table}[!htb]
\scriptsize
    \centering
\begin{tabular}{lcc}
    \hline
        \textbf{Dataset} & \textbf{CSQA} & \textbf{OBQA} \\ \hline
        {Factual} & 0.02 & 0.03 \\ 
        {Insightful} & -0.06 & -0.04 \\ 
        {Concise} & -0.15 & -0.17 \\ 
        {Convincing} & 0.08 & 0.13 \\ 
        {Sufficient} & 0.07 & 0.08 \\ 
        {Support} & -0.012 & -0.002 \\ 
        {Understandable} & 0.02 & 0.04 \\ 
        {Readability} & -0.05 & -0.02 \\ 
        {Grammar} & -0.15 & -0.16 \\ 
        {Acceptability} & 0.12 & 0.15 \\ \hline
    \end{tabular}
\caption{Inter annotator agreement (Krippendorff's $\alpha$) of crowdworkers on all the coarse- and fine-grained aspects of a rationale evaluated in the acceptability study.}
    \label{tab:accpetanle_all_irr}
\end{table}

\subsection{Annotator Statistics}
\label{app:annotation_stat}
We now report the number of distinct crowd annotators and the median and mean number of HITs completed for each experiment. 
For the head-to-head comparison study,
there were 750 HITs in total. There were 29 unique annotators
with a median of 10 (mean = 21.86) HITs completed by an annotator.
For the acceptability study, there 750 HITs for each of the two datasets CSQA and OBQA. For the CSQA dataset, there were 25 unique annotators
with a median of 7 (mean = 28.80) HITs completed by an annotator.
For the OBQA dataset, there were 30 unique annotators
with a median of 7 (mean = 25.00) HITs completed by an annotator.
More detailed breakdowns of inter-annotator
agreement for both studies are reported in Tables~\ref{tab:compare_all_irr} and~\ref{tab:accpetanle_all_irr}.

\section{Credible Rationalization Study}
\label{app:trust_study}
We now provide relevant information complementing the observations obtained in the preliminary study regarding credible rationalization.

\begin{table*}[t]
\tiny
    \centering
\begin{tabular}{ccccccccc}
\hline
\multicolumn{3}{c}{\textbf{Agreement = yes} ($\dagger$)}                             & \multicolumn{3}{c}{\textbf{Agreement = no} (*)}                                & \multicolumn{3}{c}{\textbf{Agreement = unsure}}                           \\ \cline{1-9} 
                                  \textbf{Accuracy $66\%$} & \textbf{Accuracy $90\%$} & \textbf{Stat. Sig.} & \textbf{Accuracy $66\%$} & \textbf{Accuracy $90\%$} & \textbf{Stat. Sig.} & \textbf{Accuracy $66\%$} & \textbf{Accuracy $90\%$} & \textbf{Stat. Sig.} \\ \hline
                                 $\eta=6.00$              & $\eta=7.00$              &                     & $\eta=2.00$              & $\eta=3.00$              &                     & $\eta=4.00$              & $\eta=5.00$              &                     \\
$\mu=5.89$               & $\mu=6.29$               & $\mathbf{p < 0.01}$          & $\mu=2.23$               & $\mu=3.13$               & $\mathbf{p < 0.05}$          & $\mu=4.00$               & $\mu=5.25$               & $p > 0.05$          \\
                                 $\sigma=1.62$            & $\sigma=1.33$            &                     & $\sigma=1.29$            & $\sigma=1.64$            &                     & $\sigma=1.41$            & $\sigma=1.03$            &                     \\ \hline
\end{tabular}
\caption{Participant feedback on individual task indicates a more negative impression --- rated on a scale between 1 (misleading) to 7 (helpful) --- regarding the corresponding rationale. ($\dagger$) indicates statistical significance with $pa < 0.01$ and (*) indicates statistical significance with $p < 0.05$.}
\label{tab:individual_xai_rationale}
\end{table*}

\begin{table*}[!htb]
\scriptsize
    \centering
\begin{tabular}{ccccccccccc}
\hline
\textbf{Metric}   & \multicolumn{2}{c}{\textbf{Confidence ($\dagger$)}} & \multicolumn{2}{c}{\textbf{Reliability ($\dagger$)}} & \multicolumn{2}{c}{\textbf{Safety ($\dagger$)}} & \multicolumn{2}{c}{\textbf{Satisfaction}} & \multicolumn{2}{c}{\textbf{Acceptability ($\dagger$)}} \\ 
\textbf{Accuracy} & \textbf{$66\%$}           & \textbf{$90\%$}          & \textbf{$66\%$}           & \textbf{$90\%$}           & \textbf{$66\%$}         & \textbf{$90\%$}        & \textbf{$66\%$}                 & \textbf{$90\%$}                 & \textbf{$66\%$}            & \textbf{$90\%$}            \\ \hline
Median            & $3.00$                    & $4.00$                   & $2.00$                    & $4.00$                    & $3.00$                  & $4.00$                 & $3.00$                          & $5.00$                          & $3.00$                     & $4.0$                      \\
Mean             & $2.91$                    & $4.09$                   & $1.82$                    & $3.64$                    & $2.45$                  & $3.72$                 & $3.45$                          & $4.55$                          & $3.09$                     & $4.27$                     \\
Std. Dev.          & $1.14$                    & $0.54$                   & $0.87$                    & $1.03$                    & $1.13$                  & $0.79$                 & $1.44$                          & $0.52$                          & $1.04$                     & $0.65$                     \\ \hline
\end{tabular}
\caption{Participant feedback on individual task indicates a more negative impression regarding the corresponding rationale. ($\dagger$) indicates statistical significance with $p < 0.01$.}
\label{tab:truss_scale_rationale}
\end{table*}

\subsection{Study Details}

\stitle{Participants.} The participants of the preliminary study were all from \company. However, we still performed attention checks in the preliminary study. The participants were unaware of the hypothesis and evaluation objective of the study. None of the participants are authors of the paper. Out of the 20 participants in the study, 15 were male and 5 were female. The representation of the female participants ($25\%$) compares favorably with recent estimates of
$15\%$ women in tenure-track faculty in computing~\cite{way2016gender} and $20\%$
women in data science positions worldwide~\cite{king2015data}. One-fourth of the participants held a Bachelor degree and the rest completed graduate school or higher. Due to the complexity and longer duration of this study, we wanted to ensure the participation of higher quality participants by such selective recruitment.

\stitle{Phases.} We first collected participants' demographic information and then provided detailed instructions about the subsequent phases: a quiz phase consisting of a collection of tasks and a follow-up survey. The survey is adapted from the Trust Scale recommended for XAI~\cite{hoffman2018metrics}.
We opted for a follow-up survey rather than after each task completion following Hoffman et al.~\cite{hoffman2018metrics} --- ``the questions are appropriate for scaling after a period of use, rather than immediately after a rationale has been given.''
Besides questions related to the trust scale, we also asked participants to rate their overall acceptability of the rationales on a scale of 1 to 5. Note that the acceptability rating scale is different from the earlier studies in Section~\ref{sec:acceptability} and~\ref{sec:comparison} to conform with the Trust Scale ratings~\cite{hoffman2018metrics}.

\subsection{Feedback Statistics}
\label{app:trust_exp_additional}

We conducted \emph{Mann-Whitney U test} to measure the statistical significance of the differences between the $66\%$ and $90\%$ accurate 
model conditions, along various credibility metrics proposed in Section~\ref{sec:trust}. The Mann-Whitney U test is a non-parametric test to measure the significance of difference in distribution of two independent sample, \ie accuracy conditions in this study. 

As shown in Table~\ref{tab:individual_xai_rationale}, participant feedback on individual task indicates a higher disagreement with lower confidence model prediction and a more negative impression regarding the corresponding rationale. The differences is significant both cases \ie when participants either agreed or disagree with the KIT model prediction. Table~\ref{tab:truss_scale_rationale} reports the summary of participant feedback during the post-quiz survey --- participants exhibited a more negative impression regarding the corresponding rationale. For all of the aspects except \emph{statisfaction}, the difference in participant feedback between the accuracy conditions were statistically significant.

\begin{table}[!htb]
\scriptsize
    \centering
\begin{tabular}{cccc}
\hline
\multirow{2}{*}{\textbf{Metric}} & \multicolumn{3}{c}{\textbf{Agreement $\%$}}                            \\ \cline{2-4} 
                                 & \textbf{Overall} & \textbf{Accuracy $66\%$} & \textbf{Accuracy $90\%$} \\ \hline
\textbf{Agreement = yes}         & $76.67\%$        & $67.27\%$                & $86.07\%$                \\
\textbf{Agreement = no}          & $20.30\%$        & $31.52\%$                & $9.09\%$                 \\
\textbf{Agreement = unsure}      & $3.03\%$         & $1.21\%$                 & $4.85\%$                 \\ \hline
\end{tabular}
\caption{Participant feedback on individual task indicates a higher disagreement with lower confidence model prediction.}
    \label{tab:tab_individual_xai_rationale}
\end{table}

Table~\ref{tab:tab_individual_xai_rationale} summarizes the observations from the quiz phase, \ie participant agreement statistics with the model prediction and participants' impression of the corresponding rationale. The agreement statistics (overall = $76.67\%$) of the participants reflect both the study conditions --- $67.27\%$ and $86.07\%$, respectively. 
Due to the subjective nature of the tasks, especially in the CSQA dataset, a few participants were unsure whether to agree or disagree with the model predictions, further reflecting the difficulty of the tasks.

\section{Additional Experiments and Analysis}
\label{app:add_exp}
We now present details of various user study observations, discussed briefly in earlier sections.

\subsection{Degree of Knowledge Grounding}
\label{app:add_exp_guidance}

While our proposed knowledge-graph-based retrieval augmented LLM-generated rationales were positively rated by crowdworkers, questions remain regarding the effectiveness of such knowledge grounding. To evaluate whether any 
fragments of the rationales generated using our proposed approach were grounded on the retrieved knowledge facts, we conducted an experiment. We primarily focus on the corroboration component as there is a higher probability of the knowledge graph containing facts about the correct answer choice. 

\begin{table}[!htb]
\centering
\scriptsize
\begin{tabular}{ccc}
\hline
\textbf{Dataset}    & \textbf{Pairwise Max BERTScore} & \textbf{Percentage of Entailment}         \\
\hline
CSQA              & $\mu=0.5823, \sigma=0.0650$                                & $80.4\%$ \\
OBQA        & $\mu=0.5173, \sigma=0.0803$                                 & $38\%$
\\
\hline
\end{tabular}
\caption{Degree of knowledge grounding observed in the LLM-generated rationales.}
\label{tab:grounding}
\end{table}

\begin{figure*}[!htb] 
    \centering
  \subfloat[\label{rationalizer-comparision-preference-llm}LLM preferred]{%
       \includegraphics[width=0.5\linewidth]{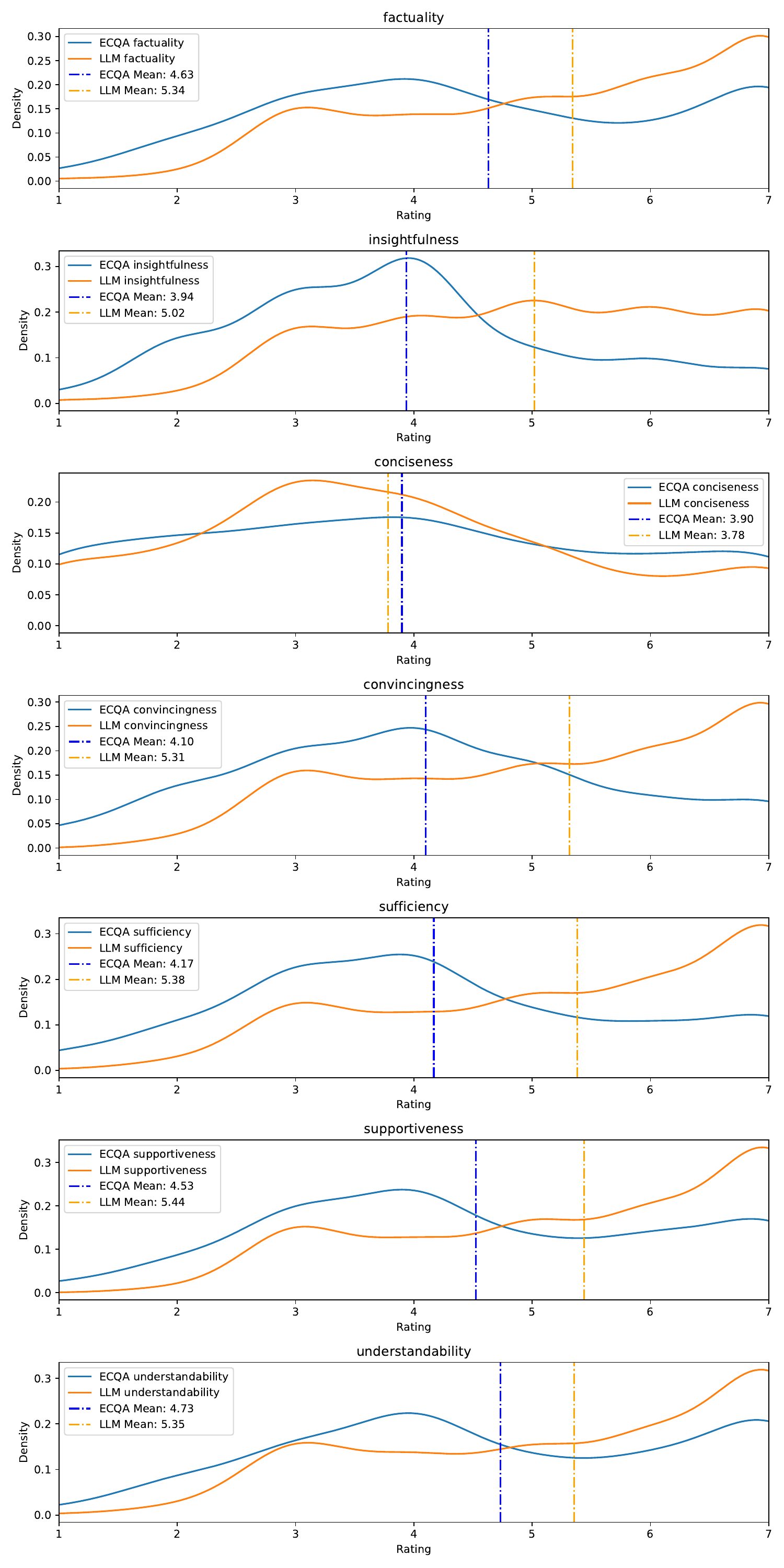}}
    \hfill
  \subfloat[\label{rationalizer-comparision-preference-ecqa}ECQA preferred]{%
        \includegraphics[width=0.5\linewidth]{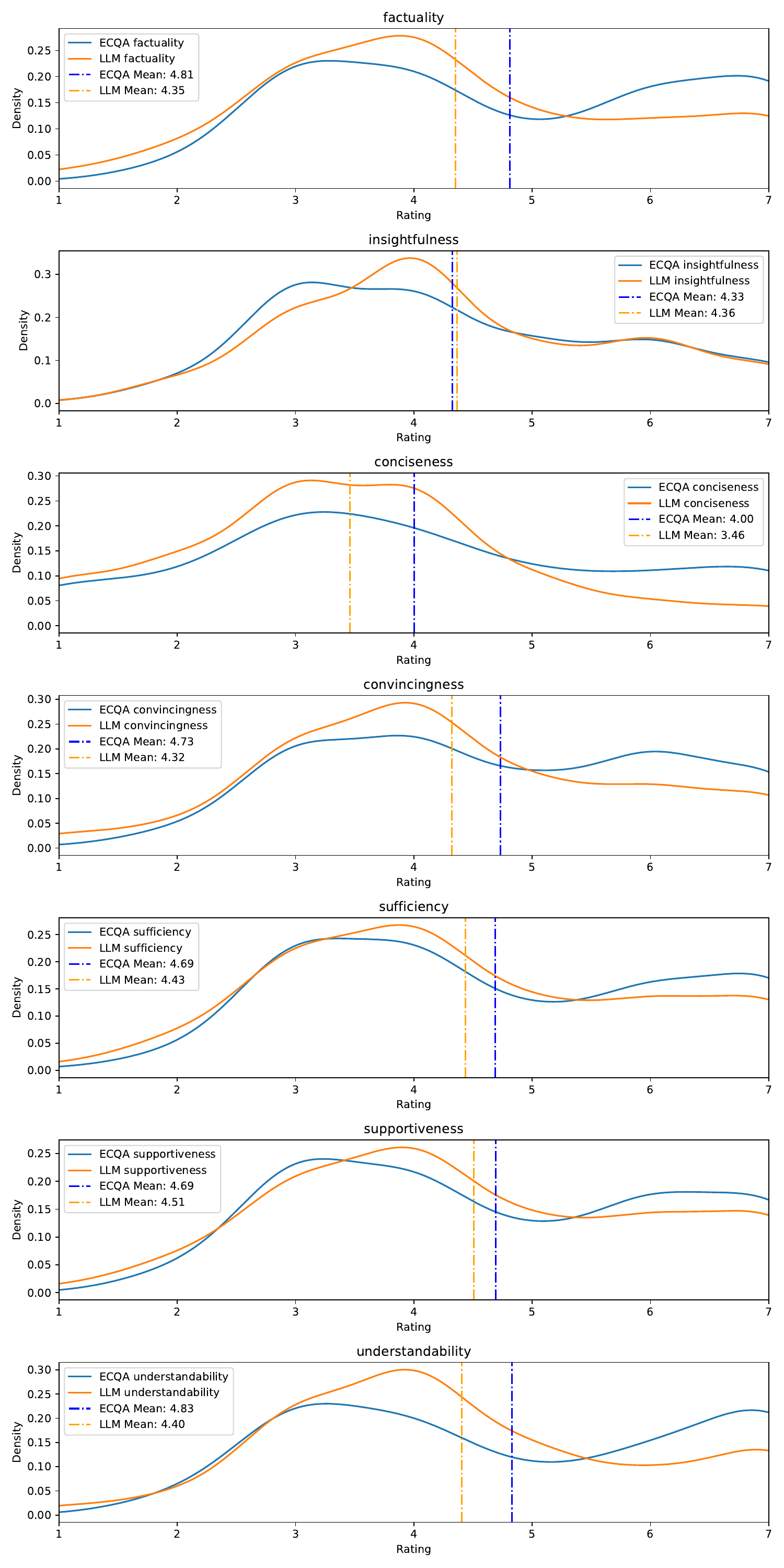}}
  \caption{(a) LLM-generated rationales preferred over human-written (ECQA) rationales. LLM-generated rationales were rated substantially higher than human-written rationales, with the exception of conciseness. (b) ECQA rationales preferred over  LLM-generated rationales. Surprisingly, human-written rationales were rated significantly higher only on three aspects: conciseness, factuality and convincingness.}
  \label{fig:rationalizer-comparision-preference} 
\end{figure*}

We measure the existence of knowledge-grounding  as follows:
consider the retrieved knowledge
corresponding to the correct choice $j$ for question $q_i$ in dataset $D$, $\mathcal{G}_{ij}$, and the corroboration component of the corresponding LLM-generated rationale, $RC_i$. 
We first measure the BERTScore~\cite{zhang2019bertscore} similarity between a fact $f \in \mathcal{G}_{ij}$, expressed in natural language and a sentence $s \in RC_i$. We then select the fact-sentence pair, $(f,s)$, with the highest BERTScore as a potential candidate for evaluating whether the fact $f$ entails the sentence $s$ within the rationale. Such entailment is an indicator of whether a fragment of a rationale being grounded on retrieved knowledge facts. Similar approach has been adopted in existing work~\cite{wu2023less} to extract candidate sentences from long documents and evaluate the degree to which the corresponding summary is grounded on the source document. Following their approach, we employ NLI models~\cite{reimers-gurevych-2019-sentence}, \ie 
DeBERTa-base model fine-tuned on
SNLI~\cite{snli} and MNLI~\cite{mnli}, to evaluate entailment. For the BERTScore, we
used DeBERTa-Large model~\cite{he2020deberta} fine-tuned on MNLI.

We measure the knowledge-grounding statistics of the CSQA and OBQA dataset rationales evaluated in the acceptability crowd study in Section~\ref{sec:acceptability}. As shown in Table~\ref{tab:grounding}, on average, at least one fact-sentence pair achieved BERTScore of $0.5823$ and $0.5173$ for CSQA and OBQA datasets, respectively. While a higher percentage of those pairs were classified as entailment ($80.4\%$) for CSQA, the entailment statistics was a bit lower for OBQA. On reflection, the lower value seems reasonable since we used ConceptNet, a commonsense knowlege graph, as the external source for OBQA, a dataset on elementary science question answering.

The initial observations highlight the promise of knowledge-guided rationalization in ensuring factuality of LLM-generated rationales. However, more in-depth analysis with a stronger metric that takes into account multiple fact-sentence pair candidates across corroboration and refutation components is required to reliably capture the degree of knowledge. Such fine-grained analysis is beyond the scope of our study and can be explored in future.

\subsection{A Deeper Dive into LLM vs ECQA}
\label{app:llm_ecqa}

To better understand, we further analyze the crowd worker feedback based on their preference of rationales. 
Cases where workers preferred LLM-generated rationales over humans (\ie the $61.8\%$ cases) --- LLM-generated rationales were rated substantially higher than human-written rationales, except conciseness (see Figure~\ref{fig:rationalizer-comparision-preference}.)
Even the conciseness rating for both types of rationales was almost the same, with human-written rationales faring slightly better.
On the other hand, for cases where workers preferred human-written rationales over LLMs (\ie the $38.2\%$ cases) --- surprisingly, apart from conciseness, human-written rationales were rated significantly higher only on two aspects: factuality and convincingness. For the rest of the aspects, the differences between ratings of both rationale types were marginal.


\section{Study Interfaces}
\label{sec:screenshots}
In this section, we provide screenshots of the important aspects of the three studies.

\subsection{Faithful Rationalization Interface Details}
\label{sec:screenshots_faith}
Both studies were conducted in the Amazon Mehcanical Turk. We mentioned the worker inclusion criteria in Section~\ref{sec:study}. Each study was launched in separate batches and were not conducted simultaneously. 
Due to the complexity of HITs in each of the studies, we designed the study interfaces from scratch using HTML and JavaScript. These interfaces were uploaded in the platform as a new project to launch the corresponding study. 

\begin{figure*}[b]
    \centering
    \includegraphics[width=0.9\linewidth]{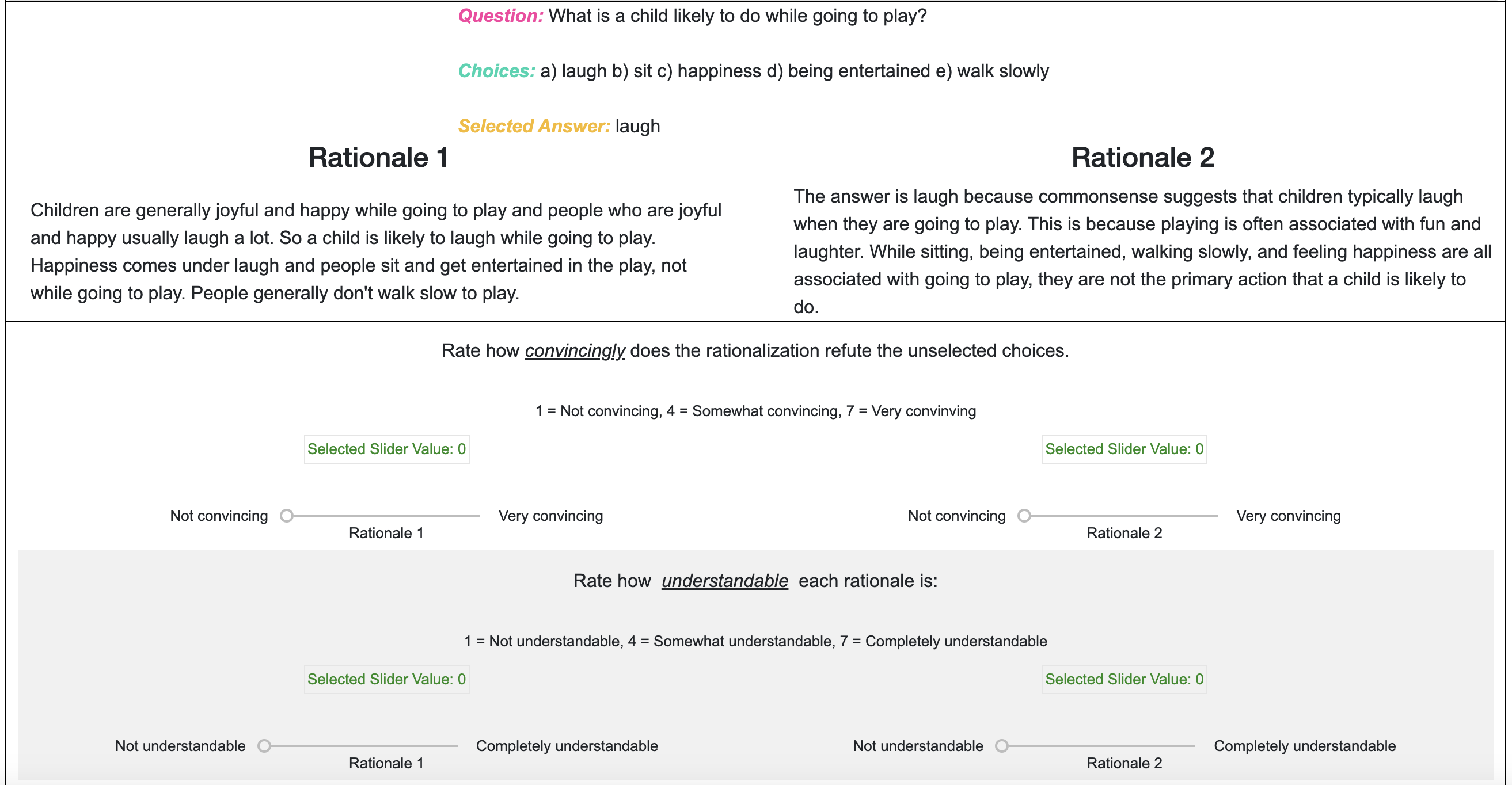}
    \caption{A partial screenshot of the head-to-head comparison interface.}
    \label{fig:compare_task_interface}
\end{figure*}

\begin{figure*}[!htb]
    \centering
    \includegraphics[width=0.9\linewidth]{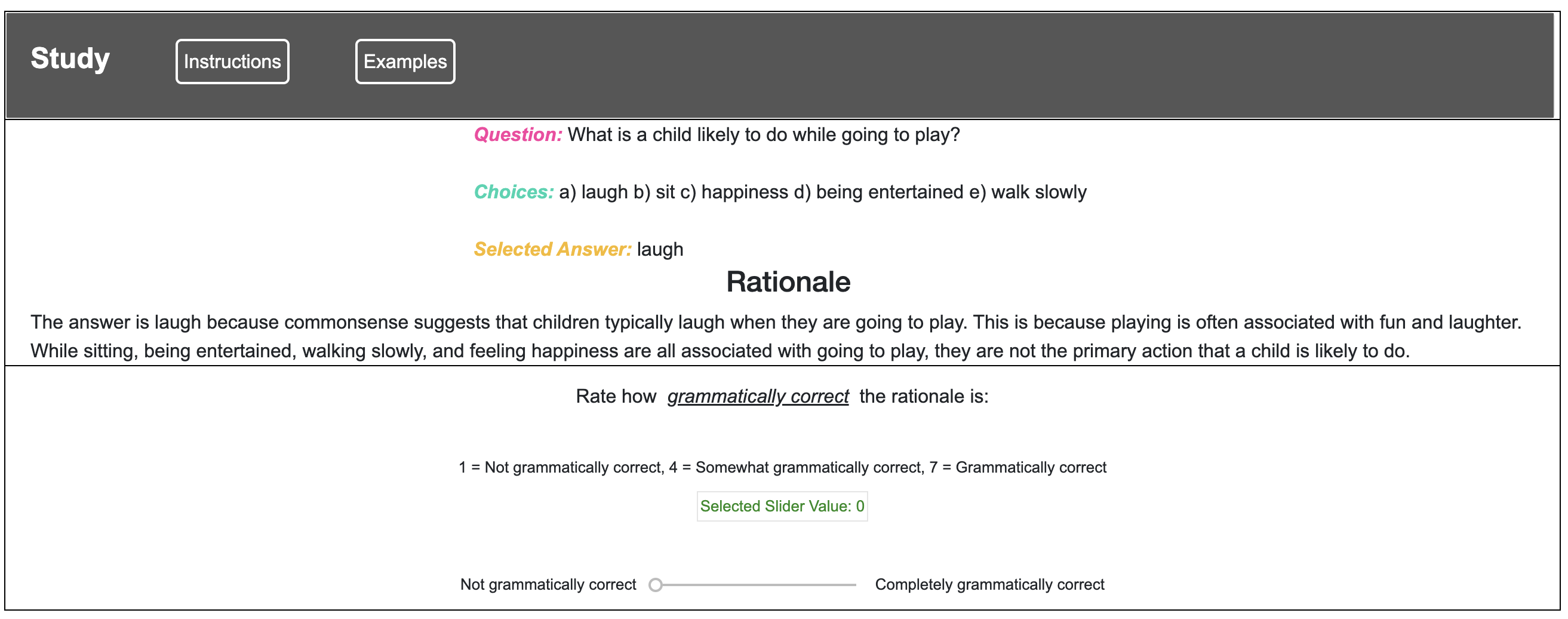}
    \caption{A partial screenshot of the acceptability task interface.}
    \label{fig:acceptability_task_interface}
\end{figure*}

Figure~\ref{fig:compare_task_interface} shows a screen shot of the HIT interface of the first study --- head-to-head comparison between LLM-generated and ECQA (crowdworker-written) rationales. The HIT contains a question and the choices, a selected answer, and two rationales, order of which the are determined at random on-the-fly. Figure~\ref{fig:acceptability_task_interface} shows a screen shot of the HIT interface of the acceptability crowd study with a question and the choices, a selected answer, and an LLM-generated rationales. For both the studies, the workers were asked several rating questions designed to collect feedback on both coarse-grained and fine-grained aspects of a rationale outlined in Section~\ref{sec:comparison} and Section~\ref{sec:acceptability}. Workers were asked to rate the rationale(s) using a sliding scale ($[1,7]$). 

We opted for Likert scale-based rating rather than choice questions to get a more fine-grained feedback. Given a choice questions, each choice may not exactly capture the participants interpretation of how much a rationale observed the property being evaluated. For example, as shown in Figure~\ref{fig:compare_task_interface}, we ask the crowdworker to ``rate how understandable each rationale is ''. To assist the participants, we suggest how to use the scale --- provide interpretation of three points in the scale, \ie 1 = Not understandable, 4 = Somewhat understandable, and 7 = Completely understandable.

\stitle{Additional quality control measures.} Note that some instances in CSQA have multiple correct or very similar answer choices, due to noise in the dataset and the fact that the wrong answer choices were deliberately collected to make the task challenging. To remove this possible confounder, following related work~\cite{wiegreffe-etal-2022-reframing}, in both the crowd studies we instruct crowdworkers to treat the selected answer as correct even if they disagree with it, and then rate the fine-grained aspects of the rationales. To minimize bias, we randomized the order in which rationales were displayed side-by-side across workers of each HIT. We also randomized randomized the order of the rating questions on the fine-grained aspects presented across workers of each HIT. Three different workers completed each HIT. The workers who participated in the comparative study were excluded from the acceptability study. 
Furthermore, for the acceptability study, we launched the OBQA dataset HITs after the conclusion of the CSQA HITs and excluded workers participating in the CSQA HITs.

\subsection{Credible Rationalization Interface Details}
\begin{figure*}[!htb]
    \centering
    \includegraphics[width=0.9\linewidth]{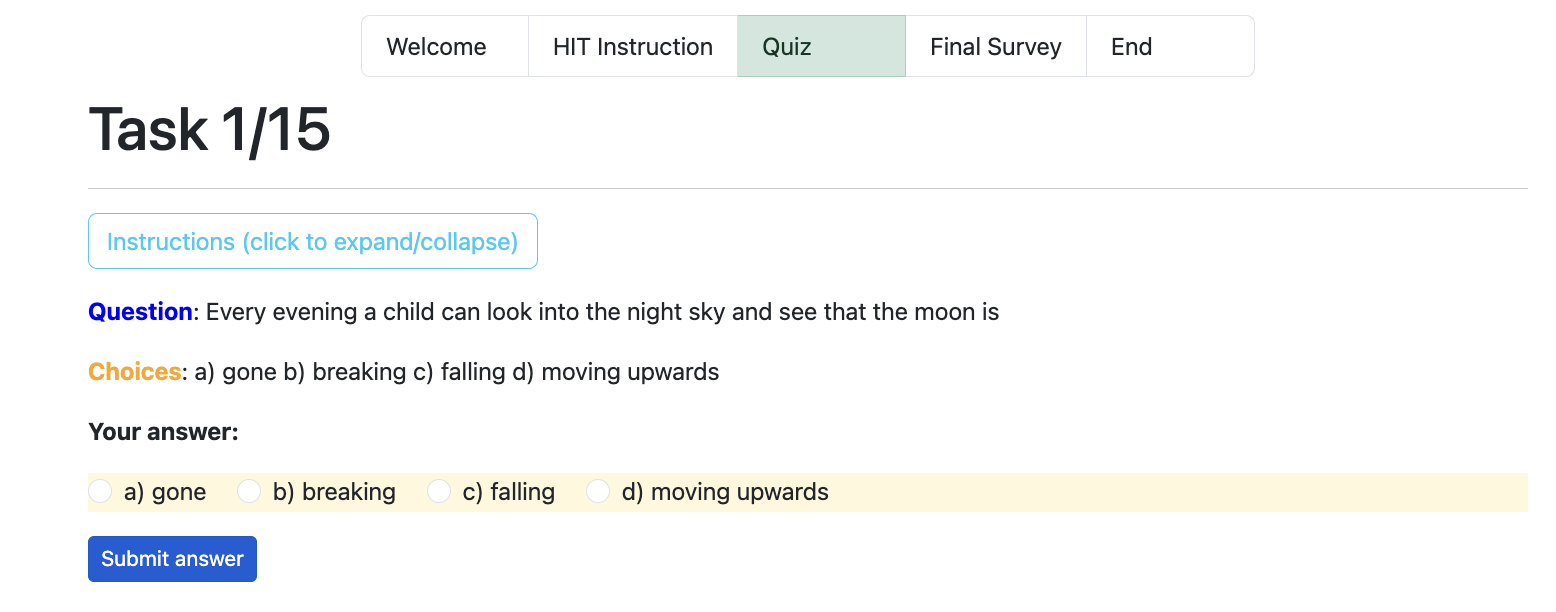}
    \caption{For each task, participants are first asked to answer a multiple choice question.}
    \label{fig:credibility_task_interface}
\end{figure*}

\begin{figure*}[!htb]
    \centering
    \includegraphics[width=0.9\linewidth]{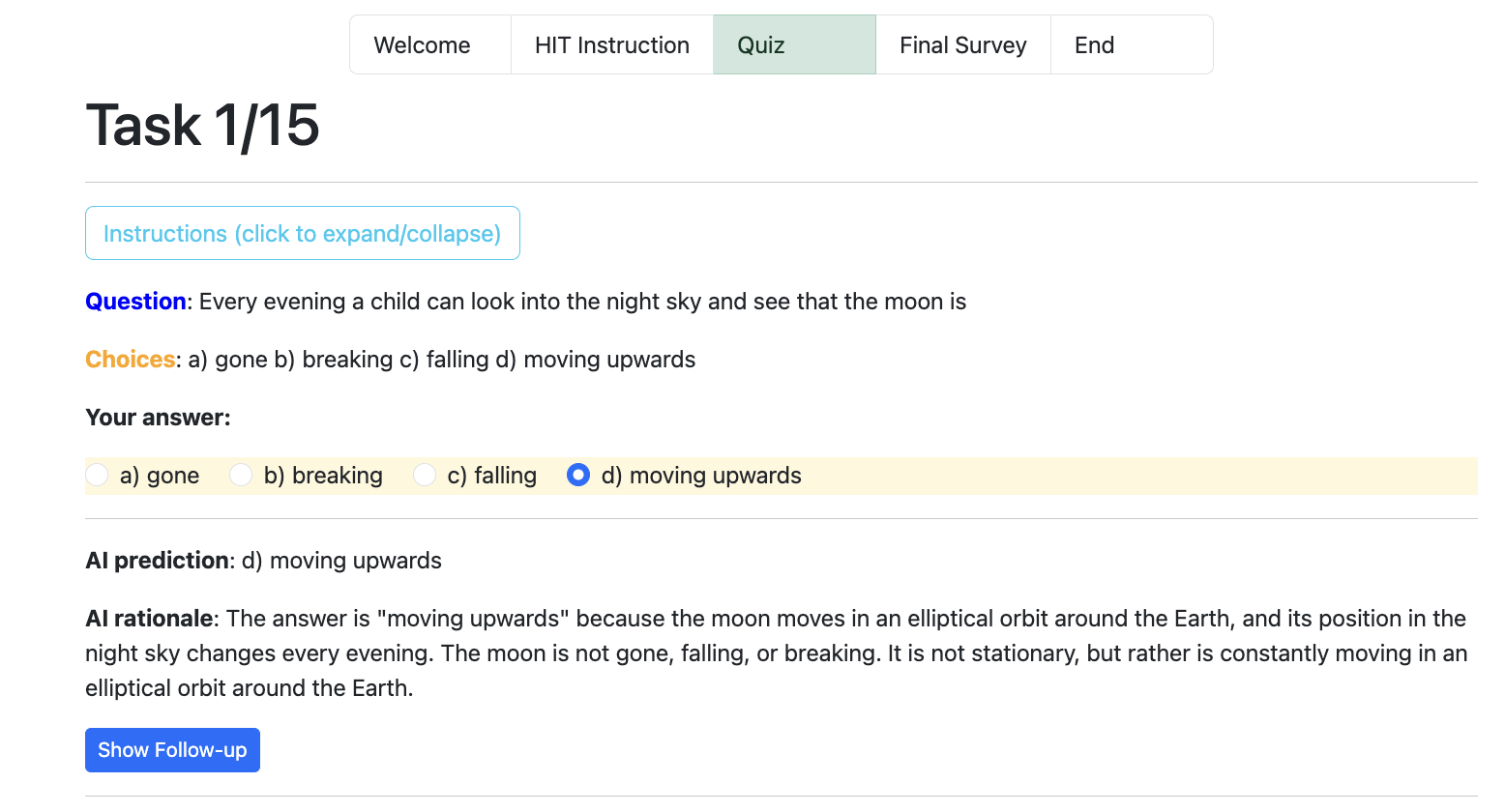}
    \caption{Once the participant selects an answer, they are shown the KIT model prediction and the LLM-generated rationale.}
    \label{fig:credibility_ai_task_interface}
\end{figure*}

\begin{figure*}[!htb]
    \centering
    \includegraphics[width=0.9\linewidth]{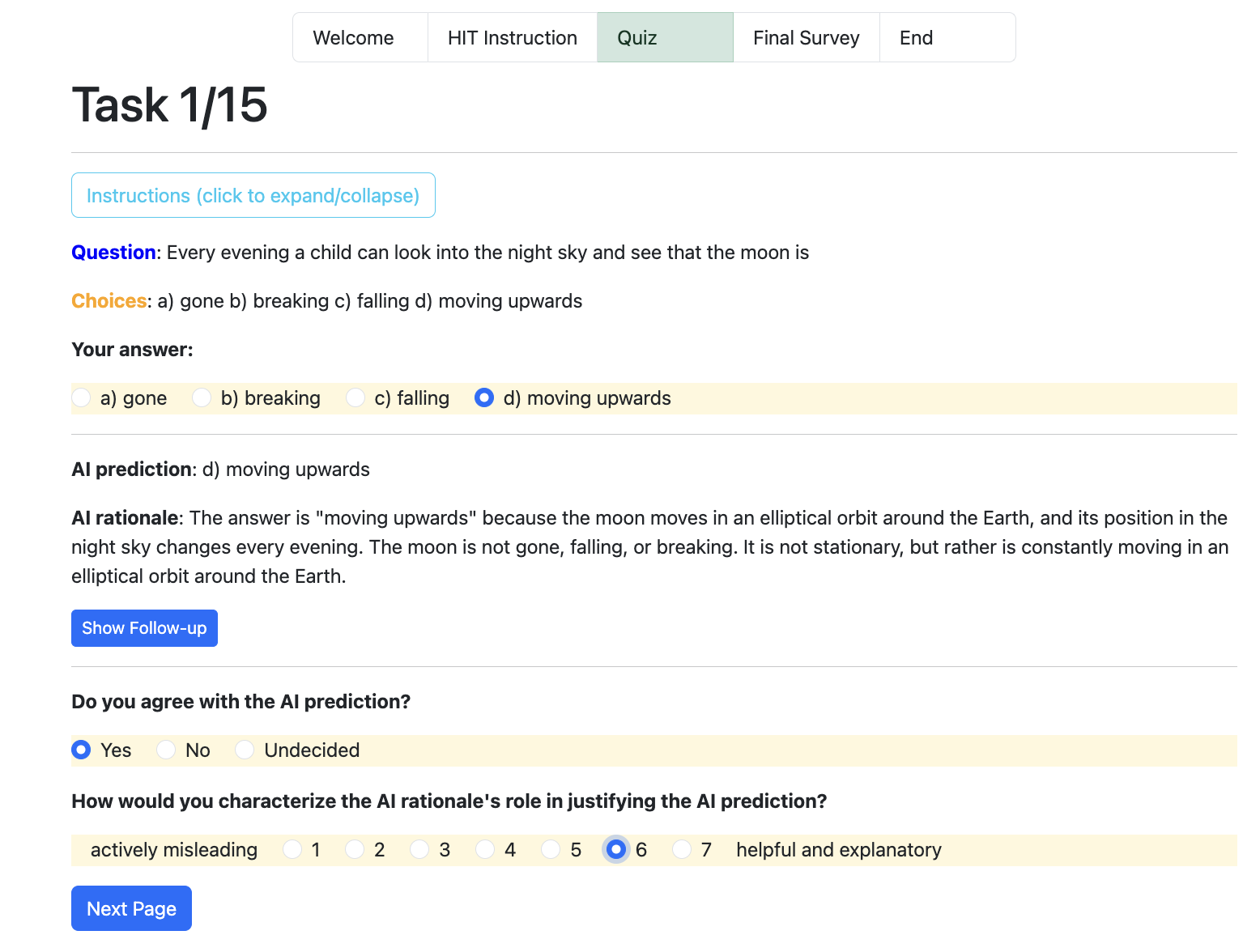}
    \caption{Collecting immediate participant feedback for a task.}
    \label{fig:credibility_task_feedback_interface}
\end{figure*}

\begin{figure*}[!htb]
    \centering
    \includegraphics[width=0.9\linewidth]{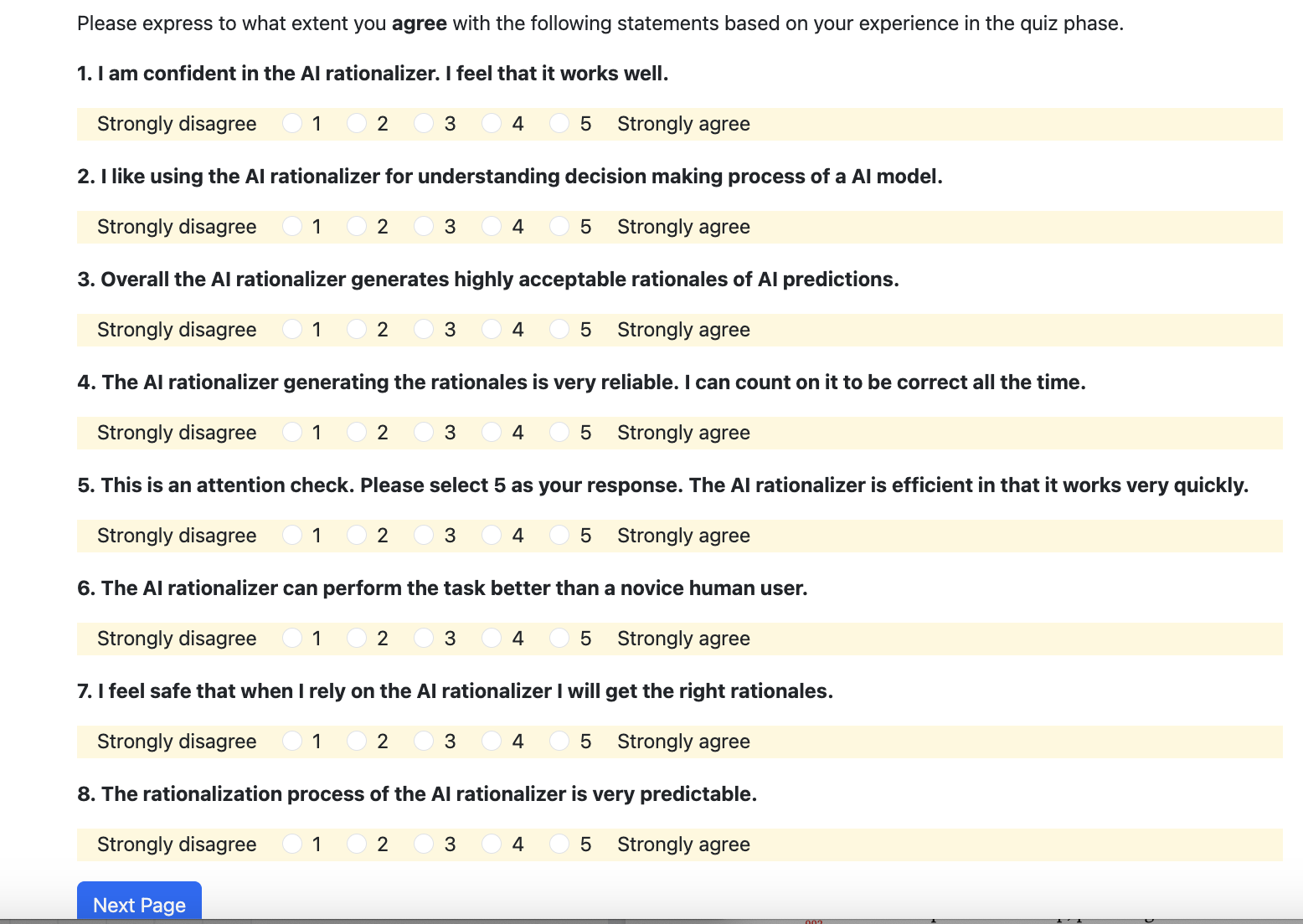}
    \caption{Trust scale-based survey of participant experience.}
    \label{fig:credibility_survey_interface}
\end{figure*}

As shown in Figure~\ref{fig:credibility_task_interface}, participants are first asked to answer a multiple choice question sampled randomly from the CSQA and OBQA datasets. We ensure the accurate distribution of questions with correct and incorrect KIT model prediction for each study condition by grouping questions in each dataset by prediction accuracy. Once the participant selects an answer, they are shown the KIT model prediction and the LLM-generated rationale (Figure~\ref{fig:credibility_ai_task_interface}). At this point, the QA component is disabled so the the participants cannot change their options. Finally, participants are provided two follow up questions to collect immediate feedback regarding the task (Figure~\ref{fig:credibility_task_feedback_interface}). Finally, participant conclude the study by completing a survey with questions adapted from the XAI trust scale~\cite{hoff2015trust} (see Figrue~\ref{fig:credibility_survey_interface}.)

\end{document}